%% file: main.tex
\documentclass[sigconf]{acmart}

\usepackage{graphicx}
\usepackage{amsmath}
\usepackage{booktabs}
\usepackage{bm}
\usepackage{multirow}
\usepackage{makecell}
\usepackage{bbding}
\usepackage{algorithmic}
\usepackage{algorithm}
\usepackage{xcolor, soul}
\usepackage{colortbl}
\usepackage{enumitem}
\usepackage{balance}
\usepackage{hyperref}
\definecolor{BLUE}{RGB}{79, 129, 189}
\definecolor{ORANGE}{RGB}{247, 150, 70}
\definecolor{GREEN}{RGB}{155, 187, 89}
\definecolor{lightBLUE}{RGB}{241, 245, 250}
\definecolor{lightORANGE}{RGB}{254, 247, 240}
\definecolor{lightGREEN}{RGB}{247, 249, 242}

\AtBeginDocument{%
  \providecommand\BibTeX{{%
    \normalfont B\kern-0.5em{\scshape i\kern-0.25em b}\kern-0.8em\TeX}}}

\copyrightyear{2023}
\acmYear{2023}
\setcopyright{acmlicensed}
\acmConference[MM '23] {Proceedings of the 31st ACM International Conference on Multimedia}{October 29--November 3, 2023}{Ottawa, ON, Canada.}
\acmBooktitle{Proceedings of the 31st ACM International Conference on Multimedia (MM '23), October 29--November 3, 2023, Ottawa, ON, Canada}
\acmPrice{15.00}
\acmISBN{979-8-4007-0108-5/23/10}
\acmDOI{10.1145/3581783.3611891}

\begin{document}

\title{CgT-GAN: CLIP-guided Text GAN for Image Captioning}

\author{Jiarui Yu}
\authornote{Both authors contributed equally to this research.}
\email{yjr@mail.ustc.edu.cn}
\orcid{0000-0001-6327-6811}
\affiliation{%
  \institution{University of Science and Technology of China}
  \city{Hefei}
  \country{China}
}

\author{Haoran Li}
\authornotemark[1]
\email{lihaoran747@126.com}
\orcid{0009-0007-3593-7169}
\affiliation{%
  \institution{University of Science and Technology of China}
  \city{Hefei}
  \country{China}
}

\author{Yanbin Hao}
\email{haoyanbin@hotmail.com}
\orcid{0000-0002-0695-1566}
\authornote{Yanbin Hao and Xiangnan He are both the corresponding authors.}
\affiliation{%
  \institution{University of Science and Technology of China}
  \city{Hefei}
  \country{China}
}

\author{Bin Zhu}
\email{andrewzhu1216@gmail.com}
\orcid{0000-0002-9213-2611}
\affiliation{%
  \institution{Singapore Management University}
  \city{Bras Basah}
  \country{Singapore}
}

\author{Tong Xu}
\email{tongxu@ustc.edu.cn}
\orcid{0000-0003-4246-5386}
\affiliation{%
  \institution{University of Science and Technology of China}
  \city{Hefei}
  \country{China}
}

\author{Xiangnan He}
\email{xiangnanhe@gmail.com}
\orcid{0000-0001-8472-7992}
\authornotemark[2]
\affiliation{%
  \institution{University of Science and Technology of China}
  \city{Hefei}
  \country{China}
}

\renewcommand{\shortauthors}{Jiarui Yu, Haoran Li, et al.}


\input{Camera_Ready/abs}



\maketitle

\input{Camera_Ready/intro_and_rel}

\input{Camera_Ready/method}
\input{Camera_Ready/exp}
\input{Camera_Ready/conclusion}
\bibliographystyle{ACM-Reference-Format}
\balance
\bibliography{Camera_Ready/ref}
\input{Camera_Ready/appendix}

\end{document}

%% file: Camera_Ready/abs.tex
\begin{abstract}
The large-scale visual-language pre-trained model, Contrastive Language-Image Pre-training (CLIP), has significantly improved image captioning for scenarios without human-annotated image-caption pairs. Recent advanced CLIP-based image captioning without human annotations follows a text-only training paradigm, \textit{i.e.}, reconstructing text from shared embedding space. Nevertheless, these approaches are limited by the training/inference gap or huge storage requirements for text embeddings. Given that it is trivial to obtain images in the real world, we propose CLIP-guided text GAN (CgT-GAN), which incorporates images into the training process to enable the model to ``see'' real visual modality. Particularly, we use adversarial training to teach CgT-GAN to mimic the phrases of an external text corpus and CLIP-based reward to provide semantic guidance. The caption generator is jointly rewarded based on the caption naturalness to human language calculated from the GAN's discriminator and the semantic guidance reward computed by the CLIP-based reward module. In addition to the cosine similarity as the semantic guidance reward (\textit{i.e.}, CLIP-cos), we further introduce a novel semantic guidance reward called CLIP-agg, which aligns the generated caption with a weighted text embedding by attentively aggregating the entire corpus. Experimental results on three subtasks (ZS-IC, In-UIC and Cross-UIC) show that CgT-GAN outperforms state-of-the-art methods significantly across all metrics. Code is available at \href{https://github.com/Lihr747/CgtGAN}{https://github.com/Lihr747/CgtGAN}.

\end{abstract}

%% file: Camera_Ready/intro_and_rel.tex
\section{Introduction}

Recently, CLIP (Contrastive Language-Image Pretraining) \cite{radford2021learning} has revolutionized the multi-modal domain by aligning images and text in a joint embedding space. CLIP has been shown to benefit numerous multi-modal tasks, such as VQA \cite{song2022clip}, text-to-image synthesis \cite{ramesh2022hierarchical}, and referring image segmentation \cite{wang2022cris}. In the studied image captioning, CLIP-based methods are also well explored in scenarios with paired training data. Prior works, such as those employing CLIP as a backbone \cite{mokady2021clipcap, barraco2022unreasonable}, or as a semantic enrichment technique \cite{li2022comprehending}, have shown improved captioning performance. Nevertheless, these methods require paired training, and the human pairwise label has a stake in the performance. In this work, we explore the feasibility of using CLIP for a more challenging task: image captioning without human-labeled pairs. That is, during training, we only utilize images and an external text corpus. The goal of the task is to generate a caption that textually describes a given image by leveraging unpaired images and sentences.

Current image captioning methods without annotations fall into two categories: concept-based and CLIP-based. Concept-based methods use computer vision techniques to discover various visual conceptual clues within images and then map the words to the caption. Existing methods \cite{feng2019unsupervised, gu2019unpaired, laina2019towards} extract objects, scenes, and attributes by using deep models pre-trained on other related tasks such as object detection \cite{huang2017speed} and scene graph generation \cite{zellers2018neural}, and advocate rewarding the generated image caption for containing the detected visual concepts. Their performances often rely heavily on the quality of visual concept extraction. Also, they are incapable of capturing complex object interactions by using such text narration that explicitly refers to visual concept \cite{gu2019unpaired}.

CLIP-based methods utilize the CLIP model as an oracle visual-language alignment tool, enabling captioning without human labels. One line of such methods \cite{tewel2022zerocap, su2022language} controls a pre-trained generative language model (LM) to produce image captions in a zero-shot manner. However, its performance is subpar due to the generative LM's poor fit for the captioning task, despite requiring no additional data. Another line of approaches \cite{gu2022can, david2022textonly, li2023decap} involves training a captioner with text-only data and reconstructing text from the CLIP text embedding. This is feasible since the text embedding shares the same cosine space with the image embedding, but these methods are often constrained by either training/inference gaps or substantial storage for textual embeddings (for projecting visual embedding to textual embedding). The final series of methods \cite{zhu2023prompt, yu2022multimodal} combine images and text corpus to compute image-text similarity using CLIP. The models are then rewarded to enhance caption grammar or identify highly correlated pseudo image-text pairs.

\begin{figure}[!t]
\centering
\includegraphics[width=0.45\textwidth]{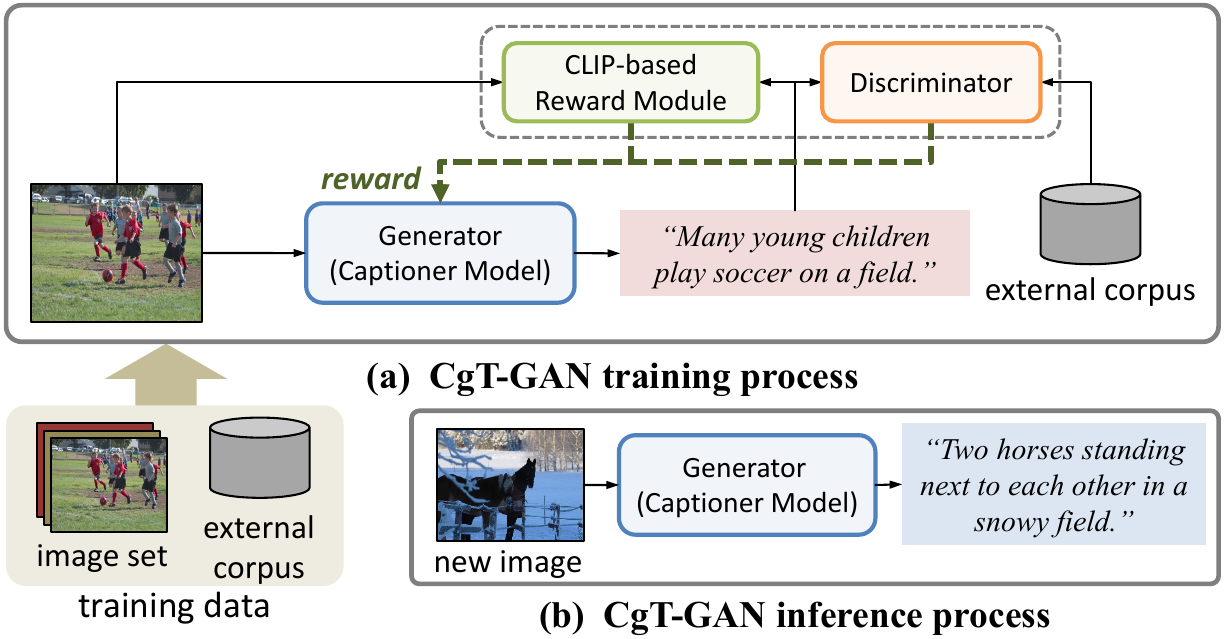}
\vspace{-0.2cm}
\caption{An illustration of our proposed CgT-GAN. (a) Rewards from the CLIP-based reward module (semantic guidance) and from the discriminator (naturalness score) are combined to guide the generator (captioner). (b) Take out the generator for new image inference.}
\vspace{-0.55cm}
\label{fig_0}
\end{figure}

In real-world applications, the images are usually easy to obtain. We, therefore, adopt the same settings as PL-UIC \cite{zhu2023prompt} and ESPER-style \cite{yu2022multimodal}, which involve the simultaneous use of images and an external corpus during training. Interestingly, even though additional images are adopted, existing solutions that use both images and sentences are inferior to text-only methods. This phenomenon motivated us to explore a more effective way of using CLIP to understand the visual modality and predict accurate captions. In this paper, as shown in Figure \ref{fig_0}, we propose a CLIP-guided text generative adversarial network (CgT-GAN), integrating CLIP \cite{radford2021learning} into the text GAN \cite{yu2017seqgan} in a more effective manner to continually guide the image-to-caption generation. Specifically, CLIP is not only for image feature encoding but also for semantic guidance rewarding. To enhance the text generation capacity of text GAN, we exploit the transformer-based language model GPT-2 \cite{radford2019language} as the generator to generate conditional synthetic caption and the improved BERT-RoBERTa \cite{liu2019roberta} as the discriminator to guess between real and fake sentences. The network is trained in an end-to-end fashion without making any extra effort on
entity detection or pseudo labelling, where the CLIP-based reward is combined with the naturalness computed by the discriminator to jointly train the generator.

Furthermore, for the reward module, we explore two rewarding strategies: CLIP-cos and CLIP-agg. CLIP-cos calculates the cosine similarity of the image-caption pair directly and rewards the generator accordingly. Inspired by \cite{li2023decap}, we additionally propose a more effective reward option, CLIP-agg, which attentively aggregates text embeddings in corpus with CLIP to guide the captioner generation.

Our contribution is three-fold:

(1) Compared to text-only CLIP-based methods, we adopt images in the training stage, which makes the captioner ``see'' real visuals to minimize the training/inference input domain gap and improve performance.

(2) Different from current CLIP-based RL or adversarial learning methods, we embed a CLIP-based reward module in a text GAN framework. Two reward strategies are proposed and analyzed for effective rewarding in an end-to-end fashion.

(3) Our model constantly outperforms the existing methods on three subtasks without human-labeled pairs: zero-shot, in-domain unpaired, and cross-domain unpaired image captioning.

\section{Related Work}
\subsection{Image Captioning}
Image captioning (IC) model learns to describe images with manually annotated image-caption pairs. Harnessing on these labels, IC training naturally focuses on maximizing the probability of correct caption. The widely used architecture of IC models is the encoder-decoder\cite{karpathy2015deep, vinyals2016show}, where the encoder captures visual content and the decoder generates caption. Riding on the structured network, attention mechanism \cite{vaswani2017attention} is also adopted to pay varying attention to image parts and tokens \cite{xu2015show, chen2017sca, lu2017knowing, anderson2018bottom, yao2018exploring, pan2020x}. Apart from model designing, advanced learning paradigms, such as reinforcement learning \cite{rennie2017self} and adversarial learning\cite{chen2019icgan, dai2017towards}, are utilized to further improve the vision-faith and text-realism for captions. 

In contrast to traditional IC, image captioning without human annotations has recently drawn researchers’ attention, where models cannot access any labelled image-text pairs. Thus, its key challenge is \textit{how to align vision and language when pairwise annotations are unavailable}. Successful attempts \cite{feng2019unsupervised, gu2019unpaired, laina2019towards} mainly follow the pipeline of teaching the caption generator to \textit{speak} human language and reducing the image-text mismatch. The goal of speaking human language is mostly achieved by utilizing either text reconstruction \cite{liu2019exploring, guo2020recurrent, meng2021object} or adversarial learning \cite{feng2019unsupervised, cao2020interactions, ben2021unpaired, song2022memorial}. Their major difference lies in the domain alignment of vision-language. Before the era of CLIP, the most representative solution is to identify features, \textit{e.g.}, visual concepts, that are common to both image and text. By harvesting visual objects or entities from images in advance, these methods can thus generate captions that contain the same visual concepts. Specifically, two kinds of concept-based alignment strategies are typically used. The first one is to learn a concept-to-sentence translator to ensure the caption being concept-related to image \cite{feng2019unsupervised,guo2020recurrent,honda2021removing,zhou2021triple,ben2021unpaired,meng2021object,zhu2022unpaired}. The other one performs concept matching in a joint embedding space, where image and text embeddings will be pulled closer if they share the same concepts, otherwise, pushed apart \cite{laina2019towards,song2022memorial}. These concept-based approaches assure the caption of containing visual concepts, nevertheless, they become ineffectual in modeling complex object correlations. To address this issue, research efforts are also made to explicitly construct scene graphs based on image objects to enrich visual context \cite{gu2019unpaired,liu2019exploring,cao2020interactions,gao2022unison}. Though providing more contextual information, the construction of a scene graph always requires complex preprocessing (\textit{e.g.}, object detection and relation formulation) and may overlook the global understanding of image content as compared to CLIP, which possesses rich vision-language knowledge.

\vspace{-0.3cm}
\subsection{CLIP for Image Captioning}
Contrastive Language–Image Pre-training~\cite{radford2021learning} (CLIP) is a large-scale multi-modal pre-training model comprised of two sub-modules: image encoder and text encoder, which aligns image and text in a joint embedding space through cosine similarity. 
CLIP has been actively exploited in many multimedia tasks\cite{patashnik2021styleclip,narasimhan2021clip-it,wang2022cris,jain2022zero,ju2022prompting,wang2023generative}. 
In the studied image captioning, the most straightforward way of using CLIP is to adopt the image encoder for more expressive feature extraction \cite{tang2021clip4caption,mokady2021clipcap,tewel2022zerocap,yu2022multimodal,zhu2023prompt}, and the text encoder for improving caption grammar \cite{cho2022fine} or semantic comprehension \cite{li2022comprehending}. 

As CLIP is trained on a web-scale image-text dataset by contrastively maximizing the visual-semantic similarity, both its image encoder and text encoder inherently possess the knowledge of cross-modal alignment. Consequently, some works \cite{tewel2022zerocap,su2022language,yu2022multimodal,zhu2023prompt,li2023decap,gu2022can, david2022textonly} consider using CLIP to teach/train a generative model for image captioning without human labelling. These works vary in the use-pattern of CLIP guidance. Specifically, \cite{tewel2022zerocap,su2022language} adjusts the language model by assessing the relatedness of each token to an image with CLIP at inference time. \cite{david2022textonly, gu2022can, li2023decap} share a similar idea of training the captioner to reconstruct text from CLIP visual-language space. These approaches only utilize text corpus during training, making them data-efficient. In detail, \cite{david2022textonly, gu2022can} address the modality gap by adding noise to the textual embeddings, while \cite{li2023decap} projects visual embedding into textual space when inference. However, the former two methods still suffer from the input difference between training and inference, and the latter requires additional textual embedding storage when projecting embedding from visual space to textual space. In contrast to these text-only methods, the proposed CgT-GAN adopts images and an external corpus during training. This arrangement allows the model to ``see'' visual modality during training, which overcomes the mentioned imperfections. Similar setting to ours, \cite{yu2022multimodal} employs a combination of CLIP reward and text likelihood reward to jointly guide the generator's learning, while \cite{zhu2023prompt} utilizes CLIP to obtain high-quality pseudo pairwise labels. Compared to these works, the CLIP-based reward module is novelly incorporated in a text GAN and provides a visual-semantic reward to guide the caption generator. Additionally, two rewarding strategies are proposed and explored in three subtasks.

%% file: Camera_Ready/method.tex
\section{METHODOLGY}
\begin{figure*}[!t]
\centering
\includegraphics[width=0.92\textwidth]{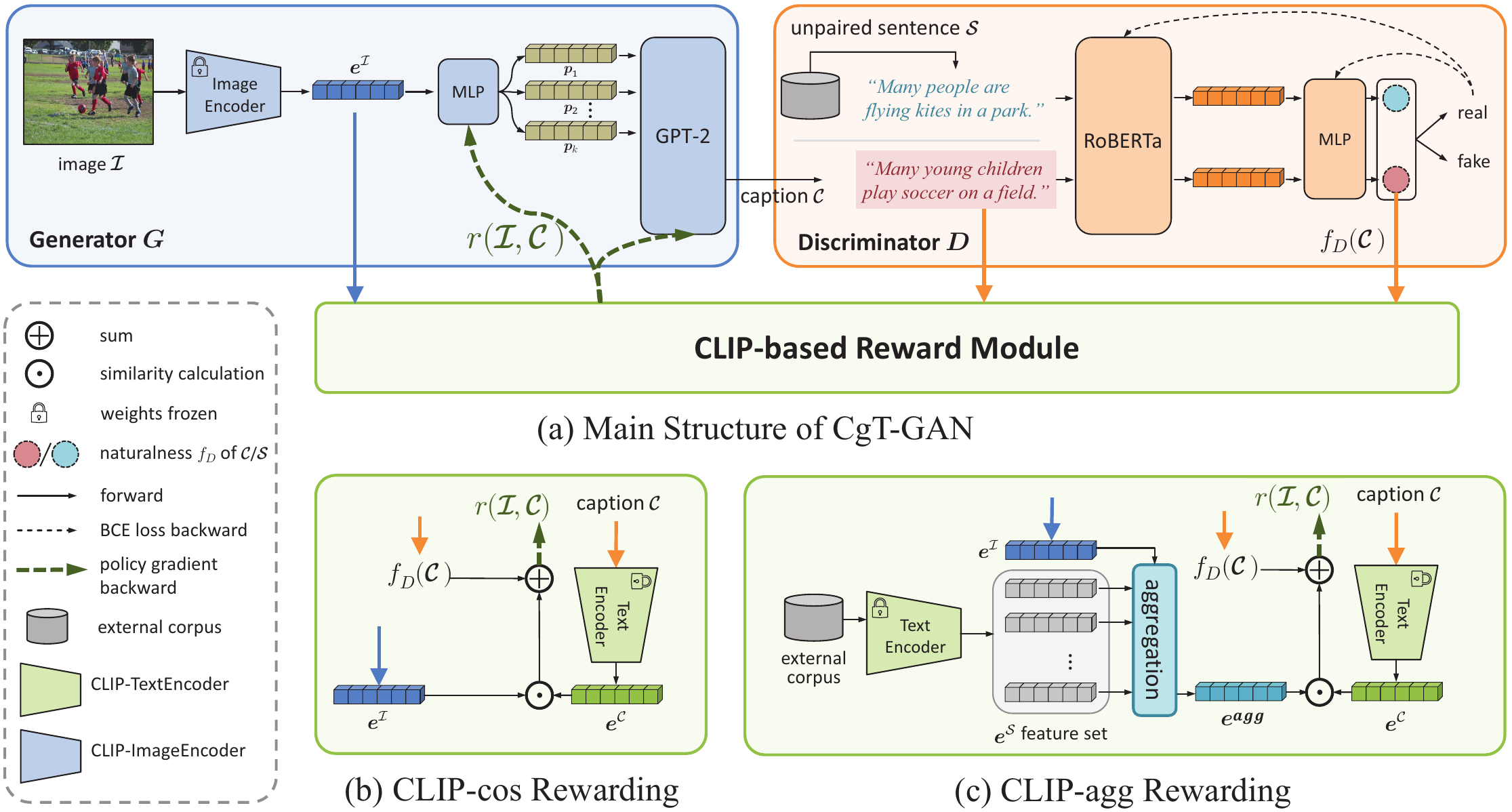}
\vspace{-0.2cm}
\caption{(a) Overall framework of CgT-GAN. CgT-GAN is composed of a text GAN module containing a caption generator $G$ ({\setlength{\fboxsep}{1pt}\fcolorbox{BLUE}{lightBLUE}{blue block}}) and a discriminator $D$ ({\setlength{\fboxsep}{1pt}\fcolorbox{ORANGE}{lightORANGE}{orange block}}) and a CLIP-based reward module ({\setlength{\fboxsep}{1pt}\fcolorbox{GREEN}{lightGREEN}{green block}}). Leveraging on the proposed CgT-GAN, the adversarial loss for real/fake sentence discrimination and a combined reward for both language naturalness and image-caption alignment are introduced. (b) and (c) depict details of the CLIP-based reward module with CLIP-cos and CLIP-agg rewarding strategies. The former computes the cosine similarity as semantic guidance, while, the latter encourages closer distance between the generated text embedding with an aggregation embedding of the corpus.
}
\vspace{-0.1cm}
\label{fig_1}
\end{figure*}

In this section, we elaborate on the details of CLIP-guided text GAN (CgT-GAN). Figure \ref{fig_1}(a) depicts the overall framework of CgT-GAN. CgT-GAN is composed of two modules, a text GAN module and a CLIP-based reward module. Instead of pure reinforcement learning or using pseudo labels by previous works \cite{zhu2023prompt, yu2022multimodal}, CgT-GAN is learned through the GAN framework, where its generator is optimized with a simple but effective reward.

\subsection{Problem Formulation} 
In this work, we focus on image captioning without using human-annotated image-caption pairs. Denote $\mathbb{I}=\{\mathcal{I}_i\}_{i=1}^{N_I}$ as a set of images, where $N_I$ indicates the total number of images. In addition, an external text corpus containing a set of sentences $\mathbb{S}=\{\mathcal{S}_i\}_{i=1}^{N_S}$, where $N_S$ denotes the total number of sentences, is generally employed to provide rich linguistic knowledge and teach the captioner to mimic human language naturally. The goal is thus to learn a mapping function $\mathcal{G}: \mathcal{I} \rightarrow \mathcal{C}$ by using $\mathbb{I}$ and $\mathbb{S}$, but without any pairwise labelling. Here, $\mathcal{C}$ refers to the generated caption.

\subsection{Text GAN Module}
\label{PG}
Similar to the vanilla GAN \cite{goodfellow2020generative} and its application variants \cite{zhu2019r2gan, liu2021aggregated}, the text GAN module in our work is also composed of a generator $G$ and a discriminator $D$. The training mechanism follows the adversarial way. 

{\bf Generator.} 
The generator $G$ is trained to generate a natural language caption $\mathcal{C}$ for an input image $\mathcal{I}$. For such an image-to-text generation task, the input image feature representation is ardently expected to contain rich language-aware information. We thus adopt the CLIP model, which well masters the vision-language prior knowledge through training on web-collected image-sentence data, to extract the visual feature. Specifically, given an image $\mathcal{I}$, we firstly obtain the feature embedding $\bm e^{\mathcal{I}} \in \mathbb{R}^{d_1}$ by using the frozen CLIP image encoder (implemented with ViT-L/14 \cite{dosovitskiy2020image}). Then, similar to ClipCap \cite{mokady2021clipcap}, a two-layer multi-layer perceptron (MLP) is employed to output a set of vectors, denoted as visual prompts, $\mathbf{P}=\left [\bm p_1, \bm p_2, \cdots, \bm p_k  \right ]$ where $\mathbf{P}\in \mathbb{R}^{k\times d_2}$. $d_1$ and $d_2$ refer to the dimension of feature and prompt embeddings. Formally, we have
\begin{equation}
\label{Image_enc}
\bm e^{\mathcal{I}}=\operatorname{CLIP-ImageEncoder}(\mathcal{I}),
\end{equation}
\begin{equation}
\label{Image_enc2}
{\bm{p}_1, \bm{p}_2, \cdots ,\bm{p}_k} = \operatorname{MLP}(\bm e^{\mathcal{I}}).
\end{equation}
Finally, the pre-trained generative language model GPT-2 \cite{radford2019language} is utilized to instantiate the caption generator $G$.  Here, GPT-2 takes the $k$ visual prompts $\{\bm p_i\}_{i=1}^{k}$ as the input tokens and continuously predicts the next word. Specifically, GPT-2 generates the $t$-th caption word $\bm{c}_t$ as:
\begin{equation}
\label{select_top}
    \bm{c}_t = \operatornamewithlimits{arg\,max}_{i} P(\bm{w}_i | \bm{p}_1, \bm{p}_2, \cdots, \bm{p}_k, \bm{c}_1, \bm{c}_2, \cdots, \bm{c}_{t-1}),
\end{equation}
where $\bm{c}_1, \bm{c}_2, \cdots, \bm{c}_{t-1}$ are prefix tokens predicted before $t$-th step, $\bm w_i$ is the $i$-th entry token in GPT-2's word dictionary and $P$ is the conditional probability. The sentence decoding stops when the sequence is as long as enough or meets the end-of-sequence (``EOS'') token. We set the max length as 20 and the ``EOS'' token as ``.''.

{\bf Discriminator.} 
The discriminator $D$ is to distinguish between real and fake (generated) sentences, \textit{i.e.}, $\mathcal{S}$ and $\mathcal{C}$.  In practice, we employ another pre-trained natural language understanding model RoBERTa followed by a two-layer MLP as our discriminator. Similar to traditional GAN, the discriminator $D$ is trained to make a judgment about how real the generated caption is, providing feedback to make the generator $G$ generate more human-like sentences. Concretely, the caption $\mathcal{C}$ and the real sentence $\mathcal{S}$ sampled from the external corpus are separately fed into RoBERTa for obtaining discriminative representation and then passed to the MLP to calculate their naturalness (a scalar value that measures how close the sentence is to natural language) \cite{dai2017towards}. The computational process is formally described as follows:
\begin{equation}
\begin{aligned}
f_{D}(\mathcal{S}) &= \operatorname{MLP}(\operatorname{RoBERTa}(\mathcal{S})), \\
f_{D}(\mathcal{C}) &= \operatorname{MLP}(\operatorname{RoBERTa}(\mathcal{C})).
\end{aligned}
\label{Naturalness}
\end{equation}

After obtaining the naturalness scores $f_{D}(\mathcal{S})$ and $f_{D}(\mathcal{C})$, the generator and discriminator can be optimized alternatively by adversarial training. However, due to the discreteness of generated caption, the gradient cannot be directly backpropagated from the discriminator to the generator. To tackle this problem, we regard the GAN learning as a reinforcement learning (RL) and use the policy gradient to train the network. The model training will be explained in detail in the section below.

\subsection{CLIP-based Reward Module}
\label{sec:reward}
The text GAN can only make the generated caption more human-like. The remaining key problem is how to align the caption with the image. In other words, the generated caption should semantically describe the image content. Recall that image-caption pairwise data is unavailable in our settings. Therefore, we propose a CLIP-based reward module to achieve image-caption semantic alignment, which produces a \textit{semantic guidance reward} to further adjust the generator $G$. For the reward module, two rewarding strategies are proposed: \textbf{CLIP-cos} and \textbf{CLIP-agg}.

\textbf{CLIP-cos.} CLIP-cos simply calculates the CLIP similarity between the image $\mathcal{I}$ and the generated caption $\mathcal{C}$, \textit{i.e.}, the cosine similarity of their embeddings. Specifically, as shown in Figure \ref{fig_1}(b), the generated caption $\mathcal{C}$ is fed into the frozen CLIP text encoder, resulting in a text embedding vector $\bm e^{\mathcal{C}}$ as:
\begin{equation}
\label{CLIPtextemb}
\bm {e}^{\mathcal{C}}=\operatorname{CLIP-TextEncoder}(\mathcal{C}).
\end{equation}
Afterwards, given the CLIP-based image embedding $\bm e^{\mathcal{I}}$ and caption embedding $\bm e^{\mathcal{C}}$, the cosine similarity can be easily calculated by 
\begin{equation}
\label{r_cos}
r_\text{cos}(\mathcal{I}, \mathcal{C}) = \cos(\bm{e}^{\mathcal{I}}, \bm{e}^{\mathcal{C}}) = \frac{\bm{e}^{\mathcal{I}} \cdot \bm{e}^{\mathcal{C}}}{\left |\bm{e}^{\mathcal{I}} \right | \left |\bm{e}^{\mathcal{C}} \right |}.
\end{equation}
We regard the cosine similarity $\cos(\bm{e}^{\mathcal{I}}, \bm{e}^{\mathcal{C}})$ as the CLIP-cos reward $r_\text{cos}$ used for the text GAN. As CLIP is pre-trained for vision-language matching by cosine score, the reward can provide robust semantic guidance.

\textbf{CLIP-agg.}  As analyzed in \cite{liang2022mind}, CLIP segregates image and text embeddings into two narrow, discrete cone-shaped spaces, known as \textit{modality gap}. This indicates that the cross-modal alignment of CLIP-cos may be inefficient. In contrast, as shown in Figure \ref{fig_1}(c), the aggregation operation of CLIP-agg enables the image embedding to be alternatively represented by a weighted sum of its associated text embeddings in the corpus. Subsequently, the CLIP-agg reward encourages a closer distance between the generated caption embedding and the aggregated textual embedding, facilitating image-text alignment within the shared CLIP text embedding space. To calculate the reward, we first compute the text embeddings $\{\bm e^{\mathcal{S}_i}\}$ of the external corpus. Then, we obtain an image-aware aggregated textual embedding $\bm e^{\text{agg}}$ through CLIP embeddings attention-weighted summation (aggregation in Figure \ref{fig_1}(c)):
\begin{equation}
\label{s_emb}
    \bm e^{\mathcal{S}_i} = \operatorname{CLIP-TextEncoder}(\mathcal{S}_i), i=1,2, \cdots, N_S,
\end{equation}
\begin{equation}
    \label{agg_emb}
    \bm e^{\text{agg}} = \sum_{i=1}^{N_S} \frac{\exp \left(\cos(\bm e^{\mathcal{S}_i}, \bm e^{\mathcal{I}}) / \tau\right)}{\sum_{k=1}^{N_S} \exp \left(\cos(\bm e^{\mathcal{S}_k}, \bm e^{\mathcal{I}}) / \tau\right)} * \bm e^{\mathcal{S}_i},
\end{equation}
where $N_S$ is the number of corpus sentences and $\tau$ is the temperature coefficient. The CLIP-agg reward $r_{\text{agg}}$ takes both cosine similarity and $L_1$ penalty between $\bm e^{\text{agg}}$ and $\bm e^\mathcal{C}$ into consideration, as they are both in the text domain. It is denoted as follows:
\begin{equation}
    \label{r_corpus}
    r_{\text{agg}}(\mathcal{I}, \mathcal{C}) = \cos(\bm e^\mathcal{C}, \bm e^{\text{agg}}) - L_1(\bm e^\mathcal{C}, \bm e^{\text{agg}}).
\end{equation}

This ensures that the generated caption is not only semantically similar to the image context ($\cos$) but also minimizes the difference in text embedding space between the generated text and the aggregation of the corpus sentences ($L_1$). Note that the CLIP encoders are fixed during training, and the aggregated text embedding $\bm e^{\text{agg}}$ can be calculated offline. Therefore, no additional computation for Eq. (\ref{s_emb}) and Eq. (\ref{agg_emb}) is incurred during the training process.

We choose CLIP-agg as the default semantic guidance rewarding strategy for our CgT-GAN. Both the two strategies and their combination are compared in the experiment section.
\subsection{Model Training}
There are two network blocks in CgT-GAN, \textit{i.e.}, the generator $G$ and discriminator $D$, that require training. Prior to adversarial training, the generator is initialized by training it to reconstruct sentences from the given corpus using their CLIP embeddings. This stage is referred to as \textit{initialization}. Afterwards, the pre-trained modules GPT-2 and RoBERTa will be fine-tuned empirically during the \textit{adversarial training} stage. The discriminator $D$ calculates the naturalness $f_{D}$ for real/fake sentences, which is optimized through binary cross-entropy loss minimization, as follows:
\begin{equation}
\label{D_training}
\begin{aligned}
\operatornamewithlimits{min}_{\varphi} & -\mathbb{E}_{\mathcal{S} \sim p_{\text{corpus}}} \left[\log  \sigma\left(f_{D_\varphi}(\mathcal{S})\right)\right] \\ &\qquad -\mathbb{E}_{\mathcal{C} \sim G_\theta} \left[\log\left(1-\sigma \left(f_ {D_\varphi}(\mathcal{C})\right)\right)\right],
\end{aligned}
\end{equation}
where $\mathcal{S}$ is the real sentence with a corpus distribution $p_{\text{corpus}}$, $\mathcal{C}$ is the generated caption sampled from $G$, $\varphi$ denotes the parameters of $D$ and $\sigma$ is the sigmoid function. 

We regard the generator training as an RL problem. In particular, our generator $G$ can be viewed as a policy, which predicts the next word (``action'') based on the current visual prompts and tokens (``state''). Therefore, the policy gradient can be approximated by using the REINFORCE algorithm \cite{sutton2018reinforcement} as follows:
{\small
\begin{equation}
\label{SCST}
\begin{aligned}
    \nabla J(\theta) = \mathbb{E}_{\mathcal{C}^s \sim G_\theta(\mathcal{C}|\mathcal{I})} & (R(\mathcal{I},\mathcal{C}^s)-R(\mathcal{I},\hat{\mathcal{C}})) \nabla \log G_\theta(\mathcal{C}^s|\mathcal{I}), \\
    \quad G_\theta(\mathcal{C}^s|\mathcal{I}) & = \prod_{t=1}^n G_\theta(\mathcal{C}^s_t | \mathcal{I}, \mathcal{C}^s_{1:t-1}),
\end{aligned}
\end{equation}
}where $\mathcal{C}^s$ is the caption sampled from the generator with each token being selected using the estimated probability $G_\theta$, $\hat{\mathcal{C}}$ is the predicted caption (\textit{i.e.}, each token is selected with the highest probability) under the inference algorithm, and $R(\cdot)$ is the reward function. The above gradient computation follows the self-critical sequence training (SCST) method \cite{rennie2017self}, which normalizes the reward utilizing the output of the generator's own test-time inference algorithm. SCST can achieve high performance on image captioning involving the test-mode inference (the baseline $R(\mathcal{I},\hat{\mathcal{C}})$) into the training promotes the training/test time consistency. 
In the context of vanilla text GANs, the reward function $R$ can be defined as the naturalness score $f_D(\mathcal{C})$ assigned by the discriminator $D$ to the generated caption, as shown in Eq. (\ref{Naturalness}). However, in our scenario, the objective of training the generator $G$ is two-fold: to produce captions that are both highly natural (\textit{i.e.}, resembling human language) and semantically consistent with the corresponding image. The overall reward function $R(\mathcal{I}, \mathcal{C})$ is thus composed of two components: the naturalness score $f_D(\mathcal{C})$, as computed by the discriminator $D$, and the semantic guidance reward $r_*$ calculated by the reward module:
\begin{equation}\label{ric}
    R(\mathcal{I}, \mathcal{C}) = f_{D}(\mathcal{C}) + r_*(\mathcal{I}, \mathcal{C}).
\end{equation}
$r_*$ can be $r_{\text{cos}}$ (Eq. (\ref{r_cos})) and $r_{\text{agg}}$ (Eq. (\ref{r_corpus})). Unless specified otherwise, we adopt $r_{\text{agg}}$ as $r_*$.

%% file: Camera_Ready/exp.tex
\section{Experiments}
In this section, we first introduce the datasets, evaluation metrics and tasks, and then make a thorough examination to answer the following four research questions:
\begin{itemize}
\item \textbf{RQ1}: How does CgT-GAN perform compared with current state-of-the-art methods?
\item \textbf{RQ2}: How does the performance of CgT-GAN vary with different reward combinations, and which rewarding strategy yields the best results?
\item \textbf{RQ3}: Can competitive performance be achieved by simpler CLIP usage or generative training on web-scale noisy pairs?
\item \textbf{RQ4}: How does CgT-GAN perform with CLIP of different scales?
\end{itemize}

\subsection{Experimental Setup}
{\bf Dataset.} We use two different image caption datasets, \textit{i.e.}, MSCOCO Caption Dataset~\cite{lin2014microsoft} and Flickr30k~\cite{plummer2015flickr30k}. MSCOCO contains 123,287 images with each image being annotated with five descriptions. Flickr30k has 31,783 images collected from Flickr website and also attaches five sentences to each image. We adopt the commonly used data split~\cite{lu2017knowing}. For external corpus, the used ShutterStock(SS)\cite{feng2019unsupervised}, Google Conceptual Captions(CC3M)\cite{sharma2018conceptual}, Flickr30k and MSCOCO training datasets contain 2.3M, 3.3M, 145k, and 557k sentences, respectively. Note that SS and CC3M are collected from the web, and Flickr30k and MSCOCO corpus are created by human labellers.

{\bf Evaluation.} Five commonly used evaluation metrics, including  BLEU-4~\cite{papineni-etal-2002-bleu}, ROUGE~\cite{lin-2004-rouge}, METEOR~\cite{banerjee-lavie-2005-meteor}, CIDEr~\cite{vedantam2015cider} and SPICE~\cite{anderson2016spice} are adopted for measuring the performance of methods from various perspectives. CLIP-based metrics like CLIP-S and refCLIP-S\cite{hessel2021clipscore} are not considered, because these scores share some similarities with our reward, thus unable to reflect real performance.

{\bf Tasks.} Following DeCap\cite{li2023decap}, we conduct experiments on three distinct tasks. (1) Zero-shot image captioning~(ZS-IC)\footnote[1]{The zero-short task is defined by DeCap\cite{li2023decap}, which chooses webly-collected corpora for use. In our experiment, we additionally access images in the training set.} (2) In-domain unpaired image captioning~(In-UIC). (3) Cross-domain unpaired image captioning~(Cross-UIC). In ZS-IC, we use sentence corpora crawled from the web, while in In-UIC and Cross-UIC, we use descriptions from an image caption training dataset as the corpus. Cross-UIC requires that the images and the sentence corpus come from different datasets, whereas In-UIC means the images and the corpora belong to
the same dataset but without pairwise information. Formally, the image captioning task is expressed in the form of \textit{X images $\leftrightarrow$ Y captions}, indicating that the images are sourced from the X dataset, while the captions are sourced from the Y dataset during the training phase. Concretely, we construct two ZS-IC tasks: \textit{MSCOCO images $\leftrightarrow$ SS captions} and \textit{MSCOCO images $\leftrightarrow$ CC3M captions}, two Cross-UIC tasks: \textit{Flickr30k images $\leftrightarrow$ MSCOCO captions} and \textit{MSCOCO images $\leftrightarrow$ Flickr30k captions} and two In-UIC tasks: \textit{MSCOCO images $\leftrightarrow$ MSCOCO captions} and \textit{Flickr30k images $\leftrightarrow$ Flickr30k captions}. It is important to note that the images in the validation and test sets are unseen during training.  Details of model optimization can be found in the supplementary material.

\subsection{Comparison with the State-of-the-Art Methods (RQ1)}

We compare CgT-GAN with various state-of-the-art (SOTA) methods, which can be categorized into two groups: concept-based methods and CLIP-based methods. The former detects objects to bridge the gap between visual and textual information. The latter aligns visuals and language with the help of CLIP. We designated a gray background in the table to distinguish CLIP-based methods, which include \textit{text-only methods} such as DeCap \cite{li2023decap}, CapDec \cite{david2022textonly}, and CLOSE \cite{gu2022can}, as well as \textit{methods that employ unpaired images and texts}, such as PL-UIC \cite{zhu2023prompt} and ESPER \cite{yu2022multimodal}. The comparisons of results are conducted separately on the aforementioned three settings: zero-shot image captioning (ZS-IC), in-domain unpaired image captioning (In-UIC), and cross-domain unpaired image captioning (Cross-UIC), following the same protocols.

\begin{table}[!t]
    \centering
    \caption{Performance comparison on the test split of the MSCOCO datasets under zero-shot captioning setting (with SS and CC3M corpus). Items in grey are CLIP-based methods.}
    \vspace{-0.2cm}
    \setlength{\tabcolsep}{1mm}{
    \begin{tabular}{cccccc}
    \hline
     Method &B.-4 &M. &R. &C. &S. \\
     \midrule[1pt]
    \multicolumn{6}{l}{{\textbf{{MSCOCO images $\leftrightarrow$ SS captions}}}} \\
    {UIC-GAN \cite{feng2019unsupervised}}        &5.6&12.4&28.7&28.6&8.1\\
    {R$^2$M \cite{guo2020recurrent}} &6.4&13.0&31.3&29.0&9.1\\
    {IGGAN \cite{cao2020interactions}}
    &6.5 &13.1 &30.5 &28.8 &8.2\\
    {TSGAN \cite{zhou2021triple}}
    &6.9 &13.0 &32.3 &28.9 &8.3\\
    \cite{feng2019unsupervised} + {Honda \textit{et al.} \cite{honda2021removing}} &7.1 &14.1	& 35.2 & 35.7 & 9.2\\
    \rowcolor{gray!20}
    {PL-UIC~\cite{zhu2023prompt}} & 10.0 & 16.2 & 35.8 & 45.8 & 11.6\\
    \rowcolor{gray!20}
    {DeCap \cite{li2023decap}} & 8.9 & 17.5 & --- &50.6 & 13.1 \\
    \rowcolor{gray!20}
    {\textbf{CgT-GAN (ours)}} & \textbf{11.1} & \textbf{19.0} & \textbf{37.2} &	\textbf{58.6} & \textbf{14.5}\\
    \hline
    \multicolumn{6}{l}{{\textbf{{MSCOCO images $\leftrightarrow$ CC3M captions}}}} \\
    {SME-Emb \cite{laina2019towards}} &6.5&12.9&35.1&22.7&---\\
    {R$^2$M \cite{guo2020recurrent}} & 8.3 &14.0&35.0&29.3&9.6\\
    {Honda \textit{et al.} \cite{honda2021removing}} &7.6 &13.5 &\textbf{37.3} &31.8 &8.4 \\
    \rowcolor{gray!20}
    {DeCap \cite{li2023decap}} & 8.8 & 16.0	& --- & 42.1 & 10.9 \\
    \rowcolor{gray!20}
    {\textbf{CgT-GAN (ours)}} &
    \textbf{10.9} & \textbf{16.9}	& 35.2 & \textbf{49.8} & \textbf{12.5}\\
    \hline
    \end{tabular}}
    \vspace{-0.1cm}
        \label{tab_ZSIC}
\end{table}

\begin{table}[!t]
    \centering
    \caption{Performance comparison on the test split of the MSCOCO and Flickr30k datasets under the cross-domain unpaired setting. Items in grey are CLIP-based methods.}
    \vspace{-0.2cm}
    \setlength{\tabcolsep}{1mm}{
    \begin{tabular}{cccccc}
    \hline
     Method &B.-4 &M. &R. &C. &S. \\
     \midrule[1pt]
    \multicolumn{6}{l}{{\textbf{{Flickr30k images $\leftrightarrow$ MSCOCO captions}}}} \\
     {SME-Emb \cite{laina2019towards}} &7.9	&13.0	&32.8 &9.9 & --- \\
     {UIC-GAN \cite{feng2019unsupervised} from \cite{ben2021unpaired}} &8.3&13.3&33.4&14.2& --- \\
     {R$^2$M \cite{guo2020recurrent}} &11.7&13.7&35.9&18.1&8.3\\
     {SCS \cite{ben2021unpaired}} & 13.0	&14.1	&37.8	&18.1 &---\\
     \rowcolor{gray!20}
     {DeCap \cite{li2023decap}} & 16.3 & 17.9 & ---  &	35.7 &	11.1 \\
     \rowcolor{gray!20}
     {CapDec \cite{david2022textonly}} & \textbf{17.3} & 18.6 & 42.7 & 35.7 & --- \\
     \rowcolor{gray!20}
    {\textbf{CgT-GAN (ours)}} &\textbf{17.3} & \textbf{19.6} & \textbf{43.9}	& \textbf{47.5} & \textbf{12.9}\\
    \hline
    \multicolumn{6}{l}{{\textbf{{MSCOCO images $\leftrightarrow$ Flickr30k captions}}}} \\
    \rowcolor{gray!20}
    {{CapDec\cite{david2022textonly}}} & 9.2 & 16.3& 36.7 & 27.3 & --- \\
    \rowcolor{gray!20}
    {{DeCap\cite{li2023decap}}} & 12.1	& 18.0 & --- & 44.4	& 10.9 \\
    \rowcolor{gray!20}
    {\textbf{CgT-GAN (ours)}} & \textbf{15.2} & \textbf{19.4}	& \textbf{40.9} & \textbf{58.7} & \textbf{13.4} \\
    \hline
    \end{tabular}}
    \vspace{-0.2cm}
        \label{tab_C-UIC}
\end{table}

{\bf Results on the ZS-IC task.} 
Table \ref{tab_ZSIC} presents the performance comparison of two zero-shot image captioning tasks. Our CgT-GAN consistently achieves the best results among all the competing methods on two ZS-IC tasks in terms of BLEU-4, METEOR, CIDEr and SPICE, outperforming other GAN methods (\textit{e.g.}, TSGAN) and the advanced CLIP-based method (\textit{e.g.}, PL-UIC and DeCap). For instance, on the \textit{MSCOCO$\leftrightarrow$SS} task, CgT-GAN obtains 19.0/58.6/14.5 METEOR/CIDEr/SPICE scores, which are 34.8\%/64.1\%/57.6\% better than the ensemble of Honda \textit{et al.}+UIC-GAN and 8.6\%/15.8\%/10.7\% better than DeCap, respectively. These results demonstrate the impressive capacity of CgT-GAN in generating lifelike image captions.

CLIP offers richer and more diverse visual-language knowledge than competing methods such as UIC-GAN, R2M, and SME-Emb, which only use category-limited visual concepts.  Compared to CLIP-based methods, our training paradigm is more efficient than PL-UIC with CLIP-filtered pseudo labels. Additionally, we observed that utilizing unlabelled image data during training significantly improves CgT-GAN's performance compared to text-only methods like DeCap. We also notice that CgT-GAN performs slightly worse than Honda \textit{et al.} on ROUGE with the CC3M corpus. This is mainly because Honda \textit{et al.} additionally utilizes pre-detected visual concepts, which are the key focus of ROUGE. In contrast, CgT-GAN is learned end-to-end without making any extra effort on entity detection. Furthermore, CgT-GAN trained with the SS corpus achieved higher scores than with the CC3M corpus because the SS corpus crawled using MSCOCO eighty keywords can provide more supportive sentences for generator guidance.

\begin{table}[!t]
    \centering
    \caption{Performance comparison on the test split of the MSCOCO and Flickr30k datasets under the in-domain unpaired setting. Items in grey are CLIP-based methods.}
    \vspace{-0.1cm}
    \setlength{\tabcolsep}{1.5mm}{
    \begin{tabular}{cccccc}
    \hline
     Method &B.-4 &M. &R. &C. &S. \\
\midrule[1pt]
    \multicolumn{6}{l}{{\textbf{{MSCOCO images $\leftrightarrow$ MSCOCO captions}}}} \\
     {Pivoting\cite{gu2018unpaired}} &5.4&13.2&---&17.7&---\\
     {SSR\cite{song2019unpaired}} &11.1&14.2&---&28.2&---\\
     {Coarse-SRE\cite{liu2019exploring}} &16.5&14.3&33.4&37.2&10.6\\
     {Fine-SRE\cite{liu2019exploring}} &19.7&17.4&41.9&49.7&13.3\\
     {UIC-GAN\cite{feng2019unsupervised}} &18.6&17.9&43.1&54.9&11.1\\
     {R$^2$M \cite{guo2020recurrent} from \cite{song2022memorial}} &16.0&17.3&39.7&48.4&11.2\\
     {TSGAN\cite{zhou2021triple}} 
     &18.9&18.2&43.3&55.2&11.3 \\
     {SME-Emb\cite{laina2019towards}} &19.3&20.2&45.0&61.8&12.9\\
     {MemGAN\cite{song2022memorial}} &20.0&19.9&45.1&63.6&12.9\\
     {IGGAN\cite{cao2020interactions}}
     &21.9 &21.1 &46.5 &64.0 &14.5\\
     {Graph-Align\cite{gu2019unpaired}} &21.5&20.9&47.2&69.5&15.0\\
     {SCS\cite{ben2021unpaired}} &22.8 &21.4&47.7&74.7&15.1\\
      {\cite{gu2019unpaired} + Fine-SRE\cite{liu2019exploring}} &21.8 &22.1 &48.4 &75.7 &16.1\\
      \rowcolor{gray!20}
      {PL-UIC~\cite{zhu2023prompt}} & 25.0 &22.6 & 49.4 &77.9 &15.2\\
      \rowcolor{gray!20}
      {ESPER-Style~\cite{yu2022multimodal}} &21.9 &21.9 &--- &78.2 &---\\
      \rowcolor{gray!20}
      {DeCap \cite{li2023decap}} & 24.7 &	25.0 & ---	& 91.2 & 18.7 \\
      \rowcolor{gray!20}
      {CapDec \cite{david2022textonly}} & 26.4 & 25.1 & 51.8 & 91.8 & --- \\
      \rowcolor{gray!20}
      {CLOSE \cite{gu2022can}} & 28.6	& 25.2 & ---	& 95.4 & 18.1\\
      \rowcolor{gray!20}
    {\textbf{CgT-GAN (ours)}} & \textbf{30.3}	& \textbf{26.9} & \textbf{54.5} &	\textbf{108.1} & \textbf{20.5}\\
    \hline
    \multicolumn{6}{l}{{\textbf{{Flickr30k images $\leftrightarrow$ Flickr30k captions}}}} \\
    {UIC-GAN~\cite{feng2019unsupervised} from \cite{ben2021unpaired}} &10.8 &14.2 &33.4 &15.4 & ---\\
     {SCS~\cite{ben2021unpaired}} &14.3 &15.6 &38.5 &20.5 & ---\\
     \rowcolor{gray!20}
     {CapDec~\cite{david2022textonly}} & 17.7 & 20.0 & 43.9 & 39.1 & ---\\
     \rowcolor{gray!20}
     {DeCap~\cite{li2023decap}} & 21.2	& 21.8 & --- & 56.7 & 15.2\\
      \rowcolor{gray!20}
    {\textbf{CgT-GAN (ours)}} &\textbf{24.1} & \textbf{22.6} & \textbf{48.2} & \textbf{64.9} & \textbf{16.1}\\
    \hline
    \end{tabular}}
        \label{tab_In-UIC}
\end{table}

{\bf Results on the Cross-UIC task}.
Note that images and the corpus are from the different datasets on the Cross-UIC setting. Table \ref{tab_C-UIC} shows the performance of the models on the image-corpora cross-domain settings of Flickr30k and MSCOCO. CgT-GAN outperforms other models on Cross-UIC task, demonstrating a similar performance trend to that on the ZS-IC task. Interestingly, with the same MSCOCO images, the use of Flickr30k corpus from human labellers only brings 0.2\% CIDEr relative improvement (58.6 $\rightarrow$ 58.7) compared to the SS corpus. This suggests that web-collected corpora may work as well as human descriptions for our model. 

{\bf Results on the In-UIC task}.
The In-UIC setting involves training images and corpus from the same dataset to test the upper bound of performance without supervision. Our experiments are conducted on Flickr30k and MSCOCO, as presented in Table \ref{tab_In-UIC}. To the best of our knowledge, our proposed CgT-GAN is the first to surpass a CIDEr score of 100 on the MSCOCO task with a vanilla CLIP backbone. CgT-GAN outperforms other reinforcement learning methods based on CLIP, such as ESPER-Style \cite{yu2022multimodal}, and CLIP text-only training methods, like CLOSE \cite{gu2022can}, which further demonstrates the advantages of our reinforcement strategy and the benefits of incorporating images in the training process. It is also certain that our CgT-GAN performs better than a group of traditional visual concept-based methods, like SCS and Graph-Align. Surprisingly, we found that our results on the MSCOCO dataset under the In-UIC setting (CIDEr score of \textbf{108.1}) are close to those of a similar generator, ClipCap\cite{mokady2021clipcap}, trained using image-caption pairs (CIDEr score of \textbf{113.1}).

\begin{table}[!t]
    \centering
    \caption{Performance changes with different rewards on MSCOCO test split under the ZS-IC (with CC3M captions) and the In-UIC (with MSCOCO captions) settings.}
    \vspace{-0.1cm}
    \setlength{\tabcolsep}{0.6mm}{
    \begin{tabular}{ccc|ccccc|ccccc}
    \hline
     \multirow{2}{*}{Init.} & \multirow{2}{*}{$f_D$} & \multirow{2}{*}{$r_{*}$} & \multicolumn{5}{c|}{MSCOCO $\leftrightarrow$ CC3M} & \multicolumn{5}{c}{MSCOCO $\leftrightarrow$ MSCOCO} \\
     \cmidrule(lr){4-8} \cmidrule(lr){9-13}
      & & &B.-4 &M. &R. &C. &S. & B.-4 &M. &R. &C. &S.\\
     \midrule[1pt]
     \resizebox{!}{0.75em}{\Checkmark} & & & 2.7 & 11.1 & 25.6 & 12.6 & 5.6 & 6.5 & 14.3 & 33.0 & 24.8 & 7.9\\
     \resizebox{!}{0.75em}{\Checkmark} & \resizebox{!}{0.75em}{\Checkmark} &  &  4.9 & 10.3 & 28.5 & 13.6 & 4.8 & 22.1 & 22.4 & 47.6 & 75.6 & 15.4\\
     \resizebox{!}{0.75em}{\Checkmark} &  & \resizebox{!}{0.75em}{\Checkmark} & 4.8 & \textbf{17.6} & 33.3 & 28.2 & \textbf{12.8} & 11.3 & 22.4 & 41.3 & 46.4 & 16.5\\
      & \resizebox{!}{0.75em}{\Checkmark} & \resizebox{!}{0.75em}{\Checkmark}& 9.8 & 16.6 & 34.6 & 47.1 & 12.3 & \textbf{30.4} & 26.3 & 54.1 & 105.8 & 20.0 \\
     \resizebox{!}{0.75em}{\Checkmark} & \resizebox{!}{0.75em}{\Checkmark} & \resizebox{!}{0.75em}{\Checkmark}&  \textbf{10.9} & 16.9 & \textbf{35.2} & \textbf{49.8} & 12.5 & 30.3 & \textbf{26.9} & \textbf{54.5} & \textbf{108.1} & \textbf{20.5}\\
    \hline
    \end{tabular}}
        \label{tab_ablation}
\end{table}

\subsection{Ablation Study (RQ2)}

The reward function $R(I, C)$ is the key of CgT-GAN, consisting of two components: the naturalness score $f_D$ and the semantic guidance reward $r_{*}$ (Eq. (\ref{ric})). In this subsection, we perform an ablation study to explore the influence of various combinations of rewards and rewarding strategies. The experiments are conducted on two distinct tasks: MSCOCO images with noisy descriptions (CC3M) and images descriptions in the same domain (MSCOCO).

\begin{table}[htbp]
    \centering
    \caption{Performance changes with different rewarding strategies on MSCOCO test split under the ZS-IC (with CC3M captions) and the In-UIC (with MSCOCO captions) settings.}
    \vspace{-0.1cm}
    \setlength{\tabcolsep}{0.6mm}{
    \begin{tabular}{c|ccccc|ccccc}
    \hline
    \multirow{2}{*}{Strategy}  & \multicolumn{5}{c|}{MSCOCO $\leftrightarrow$ CC3M} & \multicolumn{5}{c}{MSCOCO $\leftrightarrow$ MSCOCO} \\
    \cmidrule(lr){2-6} \cmidrule(lr){7-11}
      &  B.-4 &M. &R. &C. &S. &B.-4 &M. & R. &C. &S. \\
     \midrule[1pt]
     CLIP-cos & 7.9 & 16.3 & 33.1 & 39.7	& 11.4 & 22.9 & 24.7	& 49.3 & 88.9 & 18.9\\
     CLIP-agg & \textbf{10.9} & 16.9 & \textbf{35.2} & \textbf{49.8}	& 12.5 & \textbf{30.3} & \textbf{26.9} & \textbf{54.5} & \textbf{108.1} & \textbf{20.5} \\
     Reward-mix & 10.5 & \textbf{17.0} & 34.8 & 49.0 & \textbf{12.7} & 28.7 & 26.4	& 53.1 & 103.8 & 20.3\\
    \hline
    \end{tabular}}
        \label{tab_reward}
\end{table}

\textbf{Naturalness reward and semantic guidance reward.} As shown in Table \ref{tab_ablation}, the use of a single reward, either the naturalness $f_D$ or the semantic guidance reward $r_*$ results in less satisfactory performance. However, their combination significantly enhances the performance, validating our intention to jointly strengthen language naturalness and visual-language alignment. This finding confirms that both rewards achieve their intended goals and complement each other for image captioning. Moreover, the initialization stage produces a fairly good generator, enhancing adversarial training (Init. + $f_D$ + $r_*$) compared to that without init ($f_D$ + $r_*$).

\textbf{Rewarding Strategy.} In our main experiments, we selected the CLIP-agg strategy as the default option. We also tested another strategy, CLIP-cos, as described in Section \ref{sec:reward}. Additionally, we averaged the two rewards with equal weight to create a new strategy, Reward-mix. The results presented in Table \ref{tab_reward} show that the CLIP-agg strategy outperforms the other two strategies. Furthermore, we notice slower convergence and inferior performance when applying CLIP-cos, because of the modality gap\cite{liang2022mind} between visual and language modalities, which reduces the efficiency of CLIP-cos reward. While the CLIP-agg guides the generator with a textual embedding, eliminating the modality gap. Another interesting finding is that the Reward-mix strategy achieves competitive performance to CLIP-agg, even better SPICE. We speculate CLIP-cos may provide reliable guidance when the corpus is noisy or out-of-domain. For further experiments, please refer to the appendix.

\begin{figure*}[!t]
  \centering
\includegraphics[width=0.9\linewidth]{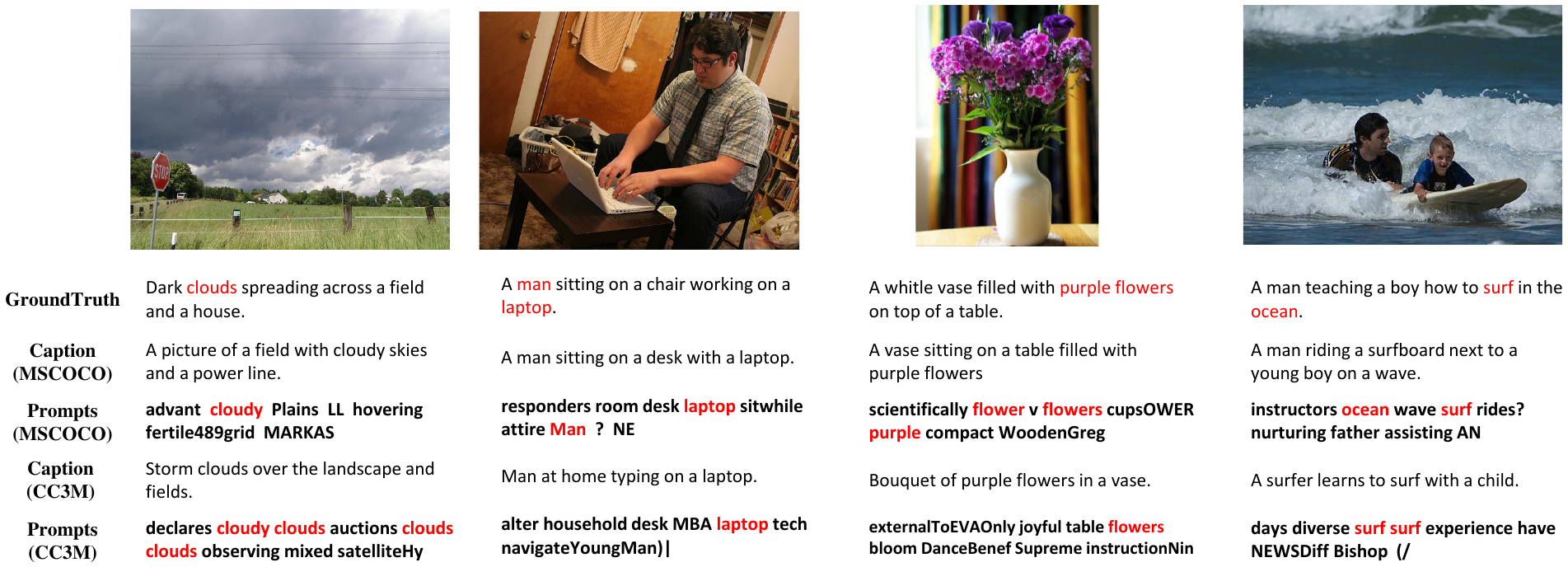}
\caption{Prompt explanations and corresponding predicted captions on MSCOCO test split. (CC3M) and (MSCOCO) denote training with CC3M and MSCOCO captions, respectively. Red fonts show that visual prompts perceive the main image content.}
\label{vis_prefix}
\end{figure*}

\subsection{Comparasion with Simpler CLIP Utilization and Web-scale Generative Training (RQ3)}
As CLIP is a powerful image-text alignment model, there are simpler methods that adopt CLIP to tackle the unpaired image captioning task. We consider two CLIP baselines: \textbf{(1) CLIP-retrieval},
where the predicted caption for each image in the test set is the corpus sentence with the highest CLIP similarity; \textbf{(2) CLIP-pseudo}, where pseudo labels are generated for each image in the training set using CLIP-retrieval on the text corpus and training is conducted using these pseudo labels. Besides, generative models pre-trained on web-collected noisy image-text pairs also seem to meet the "without human labelling" requirement. Thus, we select  \textbf{(3) SimVLM} \cite{wangsimvlm} as a comparison, a transformer-based visual language model (VLM) generatively trained on billions of web-collected image-caption pairs\cite{jia2021scaling}. Since SimVLM infers in a zero-shot manner, we compare it with CgT-GAN on ZS-IC setting.

\begin{table}[!t]\small
    \centering
    \caption{Performance comparison with generative pretraining methods and simpler CLIP-based methods using MSCOCO dataset (test set) under the ZS-IC (with CC3M captions) and the In-UIC (with MSCOCO captions) settings.}
    \vspace{-0.2cm}
    \setlength{\tabcolsep}{0.6mm}{
    \begin{tabular}{c|ccccc|ccccc}
    \hline
    \multirow{2}{*}{Method}  & \multicolumn{5}{c|}{MSCOCO $\leftrightarrow$ CC3M} & \multicolumn{5}{c}{MSCOCO $\leftrightarrow$ MSCOCO} \\
    \cmidrule(lr){2-6} \cmidrule(lr){7-11}
      &  B.-4 &M. &R. &C. &S. &B.-4 &M. & R. &C. &S. \\
     \midrule[1pt]
CLIP-retrieval & 5.5 & 13.8 & 27.7 & 31.0 & 10.0 & 13.1 & 20.8 & 40.7 & 58.0 & 15.2 \\	
CLIP-pseudo & 9.7 & 15.4 & \textbf{35.4} & 38.9 & 10.2 & 19.7 & 22.5 & 47.2 & 74.1 & 16.3 \\
SimVLM\cite{wangsimvlm} & \textbf{11.2} & 14.7 & --- & 32.2 & 8.5 & --- & --- & --- &  --- & --- \\
CgT-GAN & 10.9 &  \textbf{16.9} & 35.2 & \textbf{49.8} & \textbf{12.5} & \textbf{30.3} & \textbf{26.9} & \textbf{54.5} & \textbf{108.1} & \textbf{20.5} \\
    \hline
    \end{tabular}}
    \vspace{-0.2cm}
        \label{tab_CLIP}
\end{table}

We compare CgT-GAN with the above baselines in Table \ref{tab_CLIP}. CgT-GAN obtains much better performance than simpler CLIP-based methods: CLIP-retrieval and CLIP-pseudo. Moreover, CgT-GAN outperforms the SimVLM trained on 1.8B web-collected image-caption pairs. The comparison shows our effective usage of CLIP and unpaired data.

\subsection{CgT-GAN with Variant Backbones (RQ4)}
Current advanced CLIP-based state-of-the-arts employ CLIP encoders with varying scales, making it challenging to conduct a fair performance comparison. To ensure a fairer comparison, we conduct additional experiments to evaluate CgT-GAN with varying CLIP backbones. Table \ref{tab_backbone} summarizes the performance of CgT-GAN and CLIP-based baseline methods with different CLIP backbones. The results show that CgT-GAN performs better when CLIP backbone scales up. Moreover, our proposed CgT-GAN outperforms CLIP-based SOTAs with the same backbone.

\begin{table}[!t]
    \centering
    \caption{Performance changes with variant scale backbones using MSCOCO dataset (test set) under In-UIC setting.}
    \vspace{-0.2cm}
    \setlength{\tabcolsep}{1.0mm}{
    \begin{tabular}{c|c|ccccc}
    \hline
    Method & Backbone & B.-4 &M. &R. &C. &S. \\
     \midrule[1pt]
     ESPER\cite{yu2022multimodal} & ViT-B/32 & 21.9 & 21.9 & --- & 78.2 & --- \\
     CLOSE\cite{gu2022can} & ViT-B/32 & --- & --- & --- & 91.1 & --- \\
     DeCap\cite{li2023decap} & ViT-B/32 & 24.7 & 25.0 & --- & 91.2 & 18.7 \\
      \textbf{CgT-GAN} & ViT-B/32 & \textbf{27.4} & \textbf{25.1} & \textbf{52.0} & 	\textbf{96.9} & \textbf{18.9} \\
     \hline
     CapDec\cite{david2022textonly} & R50$\times$4 & 26.4 & 25.1 & 51.8 & 91.8 & --- \\
     CLOSE\cite{gu2022can} & R50$\times$4 & --- & --- & --- & 92.0 & --- \\
      \textbf{CgT-GAN} & R50$\times$4 & \textbf{27.2} & \textbf{25.5} & \textbf{52.3}	& \textbf{99.9} & \textbf{19.1} \\
     \hline
     CLOSE\cite{gu2022can} & ViT-L/14 & 28.6	& 25.2 & --- & 95.4 & 18.1 \\
      \textbf{CgT-GAN} & ViT-L/14 & \textbf{30.3} & \textbf{26.9} & \textbf{54.5} & \textbf{108.1} & \textbf{20.5} \\
    \hline
    \end{tabular}}
    \vspace{-0.2cm}
        \label{tab_backbone}
\end{table}

\subsection{Visual Prompts Explanation}

Visual prompts $\{\bm p_i\}_{i=1}^{k}$ are computed from the image embedding and suit for the GPT-2 sentence generation model. In other words, visual prompts are expected to work as semantic tokens to make GPT-2 generate visual-semantic consistent captions. Here, we try to understand how close the visual prompts are to the real word embeddings. For this, we compute the cosine similarity (similar to ClipCap \cite{mokady2021clipcap}) of a visual prompt and a real word embedding (from GPT-2 dictionary) and select the closest word for observation. Results are shown in figure \ref{vis_prefix}. It can be found that visual prompts surprisingly align the main image content with concept words, like ``clouds'', ``flowers'', ``surf'' and ``laptop''. Additional case studies are provided in the appendix.

%% file: Camera_Ready/conclusion.tex
\section{Conclusion}
In this paper, we have presented the CLIP-guided text GAN (CgT-GAN), which utilizes the CLIP to guide image-to-caption generation. CLIP in this paper is not only used for image encoding but also for semantic guidance. CgT-GAN allows the generator to ``see'' real images during training but does not require any human-annotated image-caption pairs. In CgT-GAN, we examine two types of CLIP-based semantic guidance rewards to enhance caption generating, including the cosine similarity reward CLIP-cos and the newly proposed text embedding aggregation reward CLIP-agg. The CLIP-based reward is finally combined with the GAN's reward to guide the generator learning in a simple and effective manner. Through extensive experiments, CgT-GAN outperforms all competing methods in three subtasks. We also showcase that the visual prompts can correspond to the salient features in the image, thereby revealing how the generator works. We want to mention here that the proposed learning fashion may also be incorporated into other GAN networks, which will be our future work.

\section{ACKNOWLEDGMENTS}
This work was supported by the National Key Research and Development Program of China under Grant 2020YFB1406703.

%% file: Camera_Ready/appendix.tex
\clearpage
\nobalance
\appendix

\section*{Appendix}

\section{Implementation Details}

We select GPT-2 model \cite{radford2019language} provided by huggingface\footnote[2]{\url{https://huggingface.co/docs/transformers/v4.21.1/en/model_doc/gpt2\#transformers.GPT2LMHeadModel}} as the generator.
The number of visual prompts is $k=10$. The MLP of the generator has 2 layers, where the channel numbers of the hidden layer and the output layer are $3840$ and $7680$, respectively. The $7680$ dimension output is reshaped to $10 \times 768$ as visual prompt vectors. For the discriminator, we employ RoBERTa-base with a pooler layer\footnote[3]{\url{https://huggingface.co/docs/transformers/model\_doc/roberta\#transformers.RobertaModel}} as our RoBERTa\cite{liu2019roberta} model. The MLP of the discriminator also has 2 layers, where the layer sizes are 384 and 1, respectively. All MLPs in our implementation take $tanh$ as the activate function. We optimize our CgT-GAN by AdamW\cite{loshchilov2018decoupled} with $\epsilon=10^{-8}$, $\beta=(0.9, 0.999)$ and weight decay $= 0.05$ on weights. The learning rate of the generator and the discriminator is set to $10^{-5}$. We set 150 warmup steps for the generator while the discriminator has no warmup steps. The batch-size is set to 128 for MSCOCO image dataset and 32 for Flickr30k image dataset. The mean of the policy gradient is estimated by sampling 5 times from the generator. For both sampling and inference, the beam size of the generator is set to 1. In practice, we train our CgT-GAN with the reward from the discriminator only for the early 150 steps, \textit{i.e.}, only using $f_D\left( C \right) $ for training. Then we linearly increase the ratio of $r_*\left( I,C \right)$ until it reaches the ratio of $f_D(C)$ in the following 2350 steps.

For the generator initialization stage, the batch-size is set to 16 for Flickr30k captions and 32 for other text corpora. The training configuration is set as: learning rate $2 \times 10 ^ {-5}$, AdamW optimizer with the same configs with GAN training, and 5000 steps warm-up to stabilize the training. 

The current implementation focuses on image captioning. We want to mention here that the CLIP (or other cross-modal alignment models)-guided GAN framework could be extended to various generative multimodal applications \cite{zhu2020cookgan, liu2021motion, 
liu2022investigating, tang2021clip4caption, jain2022zero, patashnik2021styleclip}, particularly when dealing with situations where paired data is unavailable. That will be the future work.
The CgT-GAN is trained on 2$\times$A40 GPUs, while the initialization is runing on a single A40 GPU. We use the official COCO evaluation tools\footnote[4]{\url{https://github.com/tylin/coco-caption}} to calculate all metrics.

\section{More Experimental Analysis}
\subsection{CLIP-agg Components and Temperature.} 
\begin{figure}[htbp]
  \centering
\includegraphics[width=1.00\linewidth]{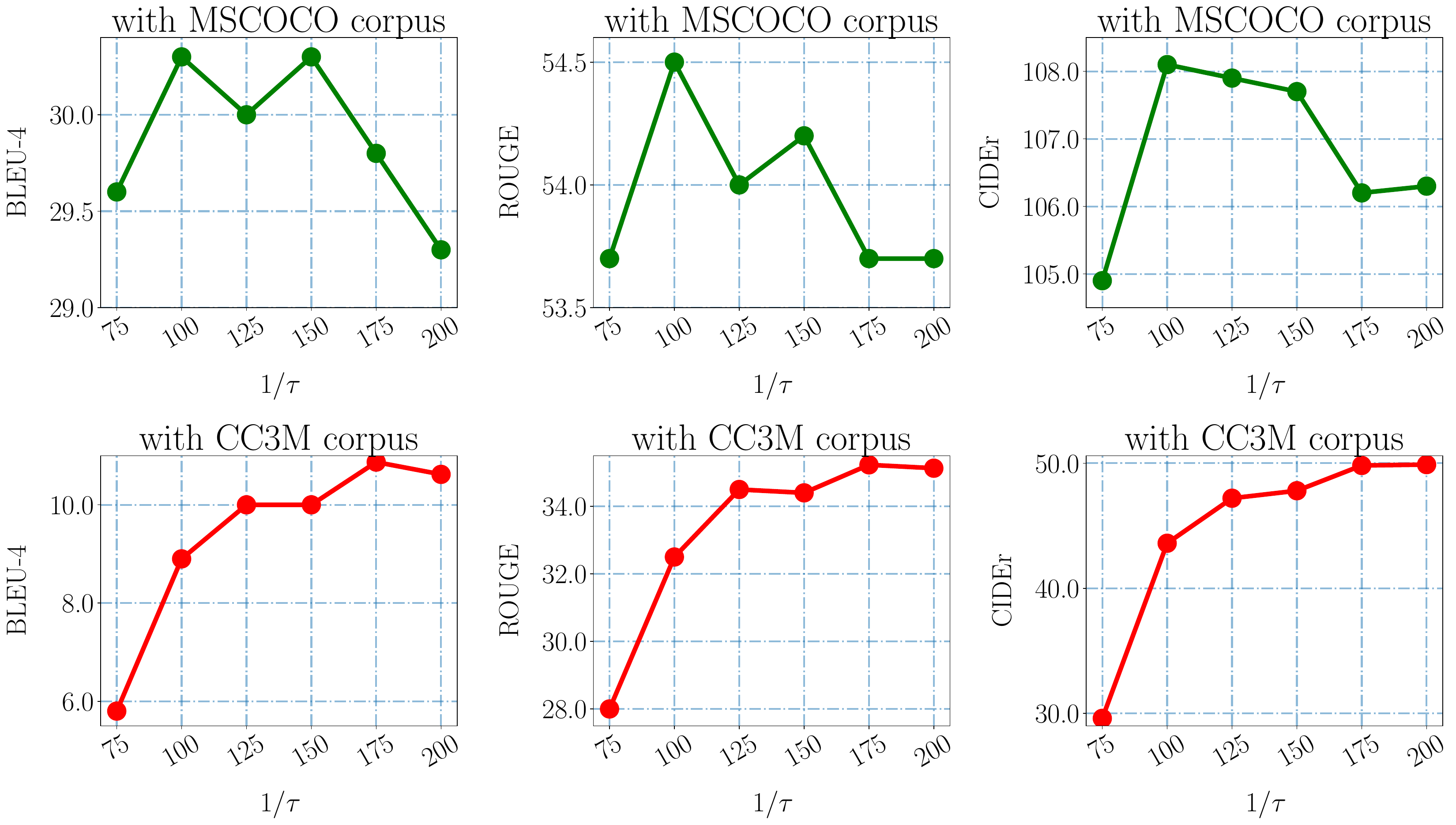}
\caption{Performance changes with different CLIP-agg temperatures on MSCOCO test split using CC3M (red lines) and MSCOCO captions (green lines).}
\label{tau}
\end{figure}

Since we set CLIP-agg as the default strategy, we further analyzed each component and temperature influence in the CLIP-agg reward. The CLIP-agg reward is composed of two components, cosine similarity and $L_1$ penalty, as expressed in Eq. (\ref{r_corpus}). Cosine similarity encourages paired embeddings to have similar semantics, consistent with the CLIP training objective. On the other hand, $L_1$ distance penalty brings caption and aggregative embeddings closer to each other in Euclidean space. Our analysis, presented in Table \ref{tab_cos_l1}, shows that the combination performs best and the two components are complementary.  The performance also varies with temperature $\tau$, as shown in Figure \ref{tau}. The temperature balances the diversity and accuracy of the supporting embeddings. A larger $\tau$ considers broader text embeddings, while a smaller $\tau$ gives more attention to closer embeddings in the aggregation. Therefore, when using a web-crawled corpus, the CLIP-agg strategy prefers a higher $\tau$ to increase the accuracy of the aggregated embedding.

\begin{table}[!t]
    \centering
    \caption{Performance changes with different CLIP-agg components on MSCOCO test split under the In-UIC setting.}
    \setlength{\tabcolsep}{1.0mm}{
    \begin{tabular}{cc|ccccc}
    \hline
    {cosine similarity}  & {$L_1$ penalty} & B.-4 &M. &R. &C. &S. \\
     \midrule[1pt]
      \resizebox{!}{0.75em}{\Checkmark} & & 30.2 & 26.6 & 53.9 & 107.3	& 20.3 \\
      & \resizebox{!}{0.75em}{\Checkmark} & 29.4 & 26.8 & 53.7 & 105.8	& \textbf{20.5} \\
     \resizebox{!}{0.75em}{\Checkmark} & \resizebox{!}{0.75em}{\Checkmark} & \textbf{30.3} & \textbf{26.9} & \textbf{54.5} & \textbf{108.1} & \textbf{20.5}\\
    \hline
    \end{tabular}}
        \label{tab_cos_l1}
\end{table}

\begin{table}[htbp]
    \centering
    \caption{Performance changes on MSCOCO test split with different corpora (CC3M and MSR-VTT) and different rewarding strategies.}
    \setlength{\tabcolsep}{0.6mm}{
    \begin{tabular}{c|ccccc|ccccc}
    \hline
    \multirow{2}{*}{Strategy}  & \multicolumn{5}{c|}{MSCOCO $\leftrightarrow$ CC3M} & \multicolumn{5}{c}{MSCOCO $\leftrightarrow$ MSR-VTT} \\
    \cmidrule(lr){2-6} \cmidrule(lr){7-11}
      &  B.-4 &M. &R. &C. &S. &B.-4 &M. & R. &C. &S. \\
     \midrule[1pt]
     CLIP-cos & 7.9 & 16.3 & 33.1 & 39.7	& 11.4 & 8.1 & \textbf{17.4}	& 36.3 & 38.2 & 11.5\\
     CLIP-agg & \textbf{10.9} & 16.9 & \textbf{35.2} & \textbf{49.8}	& 12.5 & 10.7	& 16.9 & 38.6 & 44.4 & 11.7 \\
     Reward-mix & 10.5 & \textbf{17.0} & 34.8 & 49.0 & \textbf{12.7} & \textbf{11.3} & 17.2 & \textbf{39.0} & \textbf{47.2} & \textbf{11.8} \\
    \hline
    \end{tabular}}
        \label{tab_robustness}
\end{table}

\begin{table}[!t]\small
    \centering
    \caption{Parameter count comparison with other methods. CIDEr scores are obtained under MSCOCO In-UIC setting.}
    \vspace{-0.2cm}
    \setlength{\tabcolsep}{0.5mm}{
    \begin{tabular}{c|c|cc|c}
    \hline
    Method & Encoder Version& Encoder Params. & Generator Params. & CIDEr \\
     \midrule[1pt]
     ESPER \cite{yu2022multimodal} & ViT-B/32 & 88M & 156M & 78.2 \\
      \textbf{CgT-GAN} & ViT-B/32 & 88M & 156M & 96.9 \\
     \hline
     CapDec\cite{david2022textonly} & R50$\times$4 & 87M & 182M & 91.8 \\
      \textbf{CgT-GAN} & R50$\times$4 & 87M & 156M & 99.9 \\
     \hline
     CLOSE\cite{gu2022can} & ViT-L/14 & 304M	& 225M & 95.4 \\
      \textbf{CgT-GAN} & ViT-L/14 & 304M & 157M & 108.1 \\
    \hline
    \end{tabular}}
        \label{tab_params}
\end{table}

\begin{figure*}[!t]
  \centering
\includegraphics[width=1.00\linewidth]{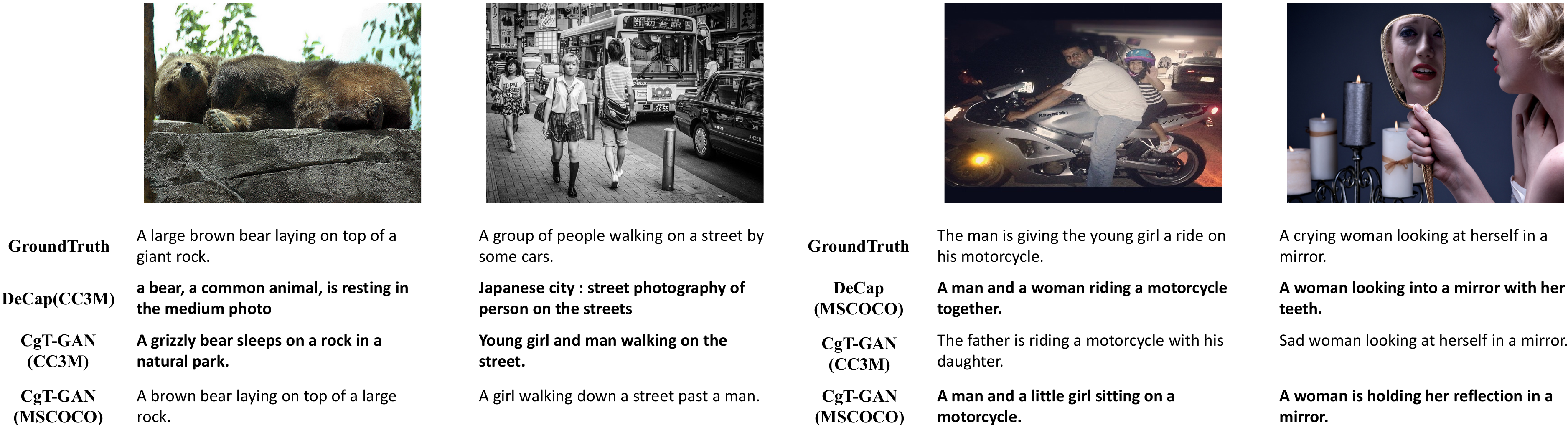}
\caption{Caption examples on MSCOCO test split of CgT-GAN. CgT-GAN (CC3M) and CgT-GAN (MSCOCO) denote training with CC3M and MSCOCO captions, respectively. Boldface fonts in the first two cases show the comparison between DeCap and our CgT-GAN with CC3M corpus. Those in the last two show comparison between CgT-GAN and Decap with MSCOCO corpus.}
\label{fig_case_study}
\end{figure*}

\begin{figure}[!t]
  \centering
\includegraphics[width=0.95\linewidth]{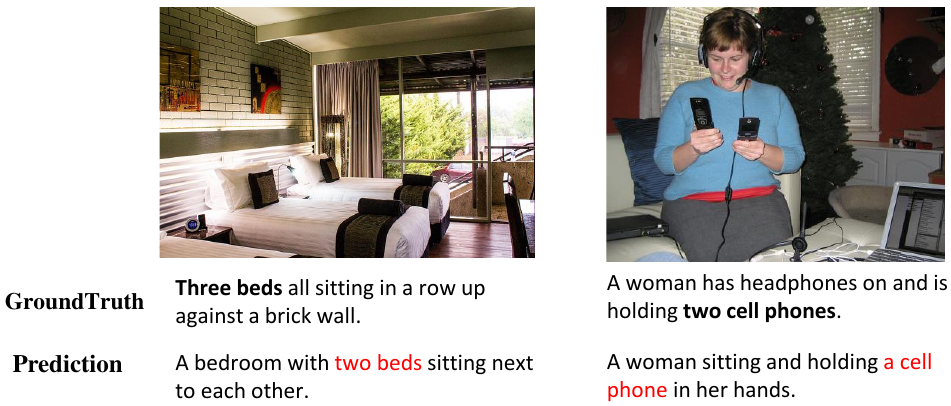}
\vspace{-0.2cm}
\caption{Failure cases on MSCOCO test split of CgT-GAN using In-UIC setting.}
\label{fail}
\end{figure}

\balance
\subsection{Robustness to Corpus Domain Variation}
During our CgT-GAN training stage, the generator is instructed to mimic the sentences in the corpus. However, there might be a large gap between the image distribution and the corpus distribution in a specific task, which has the potential to impair performance. As a result, it is important to evaluate the robustness of CgT-GAN in the face of corpus domain variations.  We conducted additional experiments using the MSCOCO images $\leftrightarrow$ MSR-VTT \cite{xu2016msr} captions setting to simulate such situations. MSR-VTT \cite{xu2016msr} is a video captioning dataset that primarily focuses on describing actions and events within the videos. Therefore, it has a significantly different caption distribution compared to the MSCOCO image dataset, which mainly consists of static visual content. Results presented in Table \ref{tab_robustness} show that CgT-GAN achieves comparable results with two diverse corpora (video captioning dataset MSR-VTT and image captioning dataset CC3M), indicating the robustness of our method in adapting to multiple corpora, even in the presence of distribution gaps. Notably, Reward-mix achieves the highest performance in the MSCOCO images $\leftrightarrow$ MSR-VTT captions setting, suggesting that the combination of CLIP-cos and CLIP-agg exhibits high robustness to the distribution gap.

\subsection{Parameter Count Comparison}
We calculate the parameters of CgT-GAN essential components in the \textit{inference stage}, i.e., the image encoder (CLIP-ImageEncoder) and the generator (MLP + GPT2). The comparison of these parameters is presented in Table \ref{tab_params}. From the results, it can be observed that our CgT-GAN has either fewer or equivalent model parameters compared to other methods, yet it achieves significantly better performance (CIDEr).

\subsection{Case Study}
We show four typical examples in Figure \ref{fig_case_study} to qualitatively compare the caption results. As can be seen, our CgT-GAN outperforms DeCap in terms of mimicking human language, especially when trained with CC3M. In the first two examples, CgT-GAN (CC3M) successfully describes the image content and produces more fluent captions than DeCap (CC3M), which indicates that adversarial learning leads to better performance than text-only methods on noisy corpora. Moreover, we discover CgT-GAN is observant to identify details and spatial relations, as shown in the last two cases, where CgT-GAN recognizes the ``little girl'' and comprehends that the ``woman'' in the mirror is a reflection. By comparing CgT-GAN (MSCOCO) and CgT-GAN (CC3M), we observe that CgT-GAN (CC3M) is more contextually imaginative, like ``natural park'' (the first case), ``daughter'' (the third case) and ``sad'' (the last case) due to the diverse training text. We also present the typical failure cases in Figure \ref{fail}, which provide insights into the potential limitation of our proposed CgT-GAN. Through the analysis of the generated captions under the In-UIC setting, it can be found that CgT-GAN encounters challenges in accurately counting objects in some cases. One possible reason is that CLIP-based visual embedding primarily focuses on high-level semantics.

%% file: main.bbl

\begin{thebibliography}{75}


\ifx \showCODEN    \undefined \def \showCODEN     #1{\unskip}     \fi
\ifx \showDOI      \undefined \def \showDOI       #1{#1}\fi
\ifx \showISBNx    \undefined \def \showISBNx     #1{\unskip}     \fi
\ifx \showISBNxiii \undefined \def \showISBNxiii  #1{\unskip}     \fi
\ifx \showISSN     \undefined \def \showISSN      #1{\unskip}     \fi
\ifx \showLCCN     \undefined \def \showLCCN      #1{\unskip}     \fi
\ifx \shownote     \undefined \def \shownote      #1{#1}          \fi
\ifx \showarticletitle \undefined \def \showarticletitle #1{#1}   \fi
\ifx \showURL      \undefined \def \showURL       {\relax}        \fi
\providecommand\bibfield[2]{#2}
\providecommand\bibinfo[2]{#2}
\providecommand\natexlab[1]{#1}
\providecommand\showeprint[2][]{arXiv:#2}

\bibitem[Anderson et~al\mbox{.}(2016)]%
        {anderson2016spice}
\bibfield{author}{\bibinfo{person}{Peter Anderson}, \bibinfo{person}{Basura
  Fernando}, \bibinfo{person}{Mark Johnson}, {and} \bibinfo{person}{Stephen
  Gould}.} \bibinfo{year}{2016}\natexlab{}.
\newblock \showarticletitle{Spice: Semantic propositional image caption
  evaluation}. In \bibinfo{booktitle}{\emph{ECCV}}. \bibinfo{pages}{382--398}.
\newblock


\bibitem[Anderson et~al\mbox{.}(2018)]%
        {anderson2018bottom}
\bibfield{author}{\bibinfo{person}{Peter Anderson}, \bibinfo{person}{Xiaodong
  He}, \bibinfo{person}{Chris Buehler}, \bibinfo{person}{Damien Teney},
  \bibinfo{person}{Mark Johnson}, \bibinfo{person}{Stephen Gould}, {and}
  \bibinfo{person}{Lei Zhang}.} \bibinfo{year}{2018}\natexlab{}.
\newblock \showarticletitle{Bottom-up and top-down attention for image
  captioning and visual question answering}. In
  \bibinfo{booktitle}{\emph{CVPR}}. \bibinfo{pages}{6077--6086}.
\newblock


\bibitem[Banerjee and Lavie(2005)]%
        {banerjee-lavie-2005-meteor}
\bibfield{author}{\bibinfo{person}{Satanjeev Banerjee} {and}
  \bibinfo{person}{Alon Lavie}.} \bibinfo{year}{2005}\natexlab{}.
\newblock \showarticletitle{{METEOR}: An Automatic Metric for {MT} Evaluation
  with Improved Correlation with Human Judgments}. In
  \bibinfo{booktitle}{\emph{ACL Workshop on Intrinsic and Extrinsic Evaluation
  Measures for Machine Translation and/or Summarization}}.
  \bibinfo{pages}{65--72}.
\newblock


\bibitem[Barraco et~al\mbox{.}(2022)]%
        {barraco2022unreasonable}
\bibfield{author}{\bibinfo{person}{Manuele Barraco}, \bibinfo{person}{Marcella
  Cornia}, \bibinfo{person}{Silvia Cascianelli}, \bibinfo{person}{Lorenzo
  Baraldi}, {and} \bibinfo{person}{Rita Cucchiara}.}
  \bibinfo{year}{2022}\natexlab{}.
\newblock \showarticletitle{The unreasonable effectiveness of CLIP features for
  image captioning: an experimental analysis}. In
  \bibinfo{booktitle}{\emph{CVPR Workshop}}. \bibinfo{pages}{4662--4670}.
\newblock


\bibitem[Ben et~al\mbox{.}(2021)]%
        {ben2021unpaired}
\bibfield{author}{\bibinfo{person}{Huixia Ben}, \bibinfo{person}{Yingwei Pan},
  \bibinfo{person}{Yehao Li}, \bibinfo{person}{Ting Yao},
  \bibinfo{person}{Richang Hong}, \bibinfo{person}{Meng Wang}, {and}
  \bibinfo{person}{Tao Mei}.} \bibinfo{year}{2021}\natexlab{}.
\newblock \showarticletitle{Unpaired image captioning with semantic-constrained
  self-learning}.
\newblock \bibinfo{journal}{\emph{TMM}}  \bibinfo{volume}{24}
  (\bibinfo{year}{2021}), \bibinfo{pages}{904--916}.
\newblock


\bibitem[Cao et~al\mbox{.}(2020)]%
        {cao2020interactions}
\bibfield{author}{\bibinfo{person}{Shan Cao}, \bibinfo{person}{Gaoyun An},
  \bibinfo{person}{Zhenxing Zheng}, {and} \bibinfo{person}{Qiuqi Ruan}.}
  \bibinfo{year}{2020}\natexlab{}.
\newblock \showarticletitle{Interactions guided generative adversarial network
  for unsupervised image captioning}.
\newblock \bibinfo{journal}{\emph{Neurocomputing}}  \bibinfo{volume}{417}
  (\bibinfo{year}{2020}), \bibinfo{pages}{419--431}.
\newblock


\bibitem[Chen et~al\mbox{.}(2019)]%
        {chen2019icgan}
\bibfield{author}{\bibinfo{person}{Chen Chen}, \bibinfo{person}{Shuai Mu},
  \bibinfo{person}{Wanpeng Xiao}, \bibinfo{person}{Zexiong Ye},
  \bibinfo{person}{Liesi Wu}, {and} \bibinfo{person}{Qi Ju}.}
  \bibinfo{year}{2019}\natexlab{}.
\newblock \showarticletitle{Improving image captioning with conditional
  generative adversarial nets}. In \bibinfo{booktitle}{\emph{AAAI}}.
  \bibinfo{pages}{8142--8150}.
\newblock


\bibitem[Chen et~al\mbox{.}(2017)]%
        {chen2017sca}
\bibfield{author}{\bibinfo{person}{Long Chen}, \bibinfo{person}{Hanwang Zhang},
  \bibinfo{person}{Jun Xiao}, \bibinfo{person}{Liqiang Nie},
  \bibinfo{person}{Jian Shao}, \bibinfo{person}{Wei Liu}, {and}
  \bibinfo{person}{Tat-Seng Chua}.} \bibinfo{year}{2017}\natexlab{}.
\newblock \showarticletitle{Sca-cnn: Spatial and channel-wise attention in
  convolutional networks for image captioning}. In
  \bibinfo{booktitle}{\emph{CVPR}}. \bibinfo{pages}{5659--5667}.
\newblock


\bibitem[Cho et~al\mbox{.}(2022)]%
        {cho2022fine}
\bibfield{author}{\bibinfo{person}{Jaemin Cho}, \bibinfo{person}{Seunghyun
  Yoon}, \bibinfo{person}{Ajinkya Kale}, \bibinfo{person}{Franck Dernoncourt},
  \bibinfo{person}{Trung Bui}, {and} \bibinfo{person}{Mohit Bansal}.}
  \bibinfo{year}{2022}\natexlab{}.
\newblock \showarticletitle{Fine-grained image captioning with clip reward}. In
  \bibinfo{booktitle}{\emph{Findings of NAACL}}. \bibinfo{pages}{517--527}.
\newblock


\bibitem[Dai et~al\mbox{.}(2017)]%
        {dai2017towards}
\bibfield{author}{\bibinfo{person}{Bo Dai}, \bibinfo{person}{Sanja Fidler},
  \bibinfo{person}{Raquel Urtasun}, {and} \bibinfo{person}{Dahua Lin}.}
  \bibinfo{year}{2017}\natexlab{}.
\newblock \showarticletitle{Towards diverse and natural image descriptions via
  a conditional gan}. In \bibinfo{booktitle}{\emph{ICCV}}.
  \bibinfo{pages}{2970--2979}.
\newblock


\bibitem[Dosovitskiy et~al\mbox{.}(2021)]%
        {dosovitskiy2020image}
\bibfield{author}{\bibinfo{person}{Alexey Dosovitskiy}, \bibinfo{person}{Lucas
  Beyer}, \bibinfo{person}{Alexander Kolesnikov}, \bibinfo{person}{Dirk
  Weissenborn}, \bibinfo{person}{Xiaohua Zhai}, \bibinfo{person}{Thomas
  Unterthiner}, \bibinfo{person}{Mostafa Dehghani}, \bibinfo{person}{Matthias
  Minderer}, \bibinfo{person}{Georg Heigold}, \bibinfo{person}{Sylvain Gelly},
  {et~al\mbox{.}}} \bibinfo{year}{2021}\natexlab{}.
\newblock \showarticletitle{An Image is Worth 16x16 Words: Transformers for
  Image Recognition at Scale}. In \bibinfo{booktitle}{\emph{ICLR}}.
\newblock


\bibitem[Feng et~al\mbox{.}(2019)]%
        {feng2019unsupervised}
\bibfield{author}{\bibinfo{person}{Yang Feng}, \bibinfo{person}{Lin Ma},
  \bibinfo{person}{Wei Liu}, {and} \bibinfo{person}{Jiebo Luo}.}
  \bibinfo{year}{2019}\natexlab{}.
\newblock \showarticletitle{Unsupervised image captioning}. In
  \bibinfo{booktitle}{\emph{CVPR}}. \bibinfo{pages}{4125--4134}.
\newblock


\bibitem[Gao et~al\mbox{.}(2022)]%
        {gao2022unison}
\bibfield{author}{\bibinfo{person}{Jiahui Gao}, \bibinfo{person}{Yi Zhou},
  \bibinfo{person}{LH Philip}, \bibinfo{person}{Shafiq Joty}, {and}
  \bibinfo{person}{Jiuxiang Gu}.} \bibinfo{year}{2022}\natexlab{}.
\newblock \showarticletitle{UNISON: Unpaired Cross-Lingual Image Captioning}.
  In \bibinfo{booktitle}{\emph{AAAI}}. \bibinfo{pages}{10654--10662}.
\newblock


\bibitem[Goodfellow et~al\mbox{.}(2020)]%
        {goodfellow2020generative}
\bibfield{author}{\bibinfo{person}{Ian Goodfellow}, \bibinfo{person}{Jean
  Pouget-Abadie}, \bibinfo{person}{Mehdi Mirza}, \bibinfo{person}{Bing Xu},
  \bibinfo{person}{David Warde-Farley}, \bibinfo{person}{Sherjil Ozair},
  \bibinfo{person}{Aaron Courville}, {and} \bibinfo{person}{Yoshua Bengio}.}
  \bibinfo{year}{2020}\natexlab{}.
\newblock \showarticletitle{Generative adversarial networks}.
\newblock \bibinfo{journal}{\emph{Commun. ACM}} \bibinfo{volume}{63},
  \bibinfo{number}{11} (\bibinfo{year}{2020}), \bibinfo{pages}{139--144}.
\newblock


\bibitem[Gu et~al\mbox{.}(2018)]%
        {gu2018unpaired}
\bibfield{author}{\bibinfo{person}{Jiuxiang Gu}, \bibinfo{person}{Shafiq Joty},
  \bibinfo{person}{Jianfei Cai}, {and} \bibinfo{person}{Gang Wang}.}
  \bibinfo{year}{2018}\natexlab{}.
\newblock \showarticletitle{Unpaired image captioning by language pivoting}. In
  \bibinfo{booktitle}{\emph{ECCV}}. \bibinfo{pages}{503--519}.
\newblock


\bibitem[Gu et~al\mbox{.}(2019)]%
        {gu2019unpaired}
\bibfield{author}{\bibinfo{person}{Jiuxiang Gu}, \bibinfo{person}{Shafiq Joty},
  \bibinfo{person}{Jianfei Cai}, \bibinfo{person}{Handong Zhao},
  \bibinfo{person}{Xu Yang}, {and} \bibinfo{person}{Gang Wang}.}
  \bibinfo{year}{2019}\natexlab{}.
\newblock \showarticletitle{Unpaired image captioning via scene graph
  alignments}. In \bibinfo{booktitle}{\emph{ICCV}}.
  \bibinfo{pages}{10323--10332}.
\newblock


\bibitem[Gu et~al\mbox{.}(2022)]%
        {gu2022can}
\bibfield{author}{\bibinfo{person}{Sophia Gu}, \bibinfo{person}{Christopher
  Clark}, {and} \bibinfo{person}{Aniruddha Kembhavi}.}
  \bibinfo{year}{2022}\natexlab{}.
\newblock \showarticletitle{I Can't Believe There's No Images! Learning Visual
  Tasks Using only Language Data}.
\newblock \bibinfo{journal}{\emph{arXiv preprint arXiv:2211.09778}}
  (\bibinfo{year}{2022}).
\newblock


\bibitem[Guo et~al\mbox{.}(2021)]%
        {guo2020recurrent}
\bibfield{author}{\bibinfo{person}{Dan Guo}, \bibinfo{person}{Yang Wang},
  \bibinfo{person}{Peipei Song}, {and} \bibinfo{person}{Meng Wang}.}
  \bibinfo{year}{2021}\natexlab{}.
\newblock \showarticletitle{Recurrent relational memory network for
  unsupervised image captioning}. In \bibinfo{booktitle}{\emph{IJCAI}}.
  \bibinfo{pages}{920--926}.
\newblock


\bibitem[Hessel et~al\mbox{.}(2021)]%
        {hessel2021clipscore}
\bibfield{author}{\bibinfo{person}{Jack Hessel}, \bibinfo{person}{Ari
  Holtzman}, \bibinfo{person}{Maxwell Forbes}, \bibinfo{person}{Ronan Le~Bras},
  {and} \bibinfo{person}{Yejin Choi}.} \bibinfo{year}{2021}\natexlab{}.
\newblock \showarticletitle{CLIPScore: A Reference-free Evaluation Metric for
  Image Captioning}. In \bibinfo{booktitle}{\emph{EMNLP}}.
  \bibinfo{pages}{7514--7528}.
\newblock


\bibitem[Honda et~al\mbox{.}(2021)]%
        {honda2021removing}
\bibfield{author}{\bibinfo{person}{Ukyo Honda}, \bibinfo{person}{Yoshitaka
  Ushiku}, \bibinfo{person}{Atsushi Hashimoto}, \bibinfo{person}{Taro
  Watanabe}, {and} \bibinfo{person}{Yuji Matsumoto}.}
  \bibinfo{year}{2021}\natexlab{}.
\newblock \showarticletitle{Removing Word-Level Spurious Alignment between
  Images and Pseudo-Captions in Unsupervised Image Captioning}. In
  \bibinfo{booktitle}{\emph{EACL}}. \bibinfo{pages}{3692--3702}.
\newblock


\bibitem[Huang et~al\mbox{.}(2017)]%
        {huang2017speed}
\bibfield{author}{\bibinfo{person}{Jonathan Huang}, \bibinfo{person}{Vivek
  Rathod}, \bibinfo{person}{Chen Sun}, \bibinfo{person}{Menglong Zhu},
  \bibinfo{person}{Anoop Korattikara}, \bibinfo{person}{Alireza Fathi},
  \bibinfo{person}{Ian Fischer}, \bibinfo{person}{Zbigniew Wojna},
  \bibinfo{person}{Yang Song}, \bibinfo{person}{Sergio Guadarrama},
  {et~al\mbox{.}}} \bibinfo{year}{2017}\natexlab{}.
\newblock \showarticletitle{Speed/accuracy trade-offs for modern convolutional
  object detectors}. In \bibinfo{booktitle}{\emph{CVPR}}.
  \bibinfo{pages}{7310--7311}.
\newblock


\bibitem[Jain et~al\mbox{.}(2022)]%
        {jain2022zero}
\bibfield{author}{\bibinfo{person}{Ajay Jain}, \bibinfo{person}{Ben
  Mildenhall}, \bibinfo{person}{Jonathan~T Barron}, \bibinfo{person}{Pieter
  Abbeel}, {and} \bibinfo{person}{Ben Poole}.} \bibinfo{year}{2022}\natexlab{}.
\newblock \showarticletitle{Zero-shot text-guided object generation with dream
  fields}. In \bibinfo{booktitle}{\emph{CVPR}}. \bibinfo{pages}{867--876}.
\newblock


\bibitem[Jia et~al\mbox{.}(2021)]%
        {jia2021scaling}
\bibfield{author}{\bibinfo{person}{Chao Jia}, \bibinfo{person}{Yinfei Yang},
  \bibinfo{person}{Ye Xia}, \bibinfo{person}{Yi-Ting Chen},
  \bibinfo{person}{Zarana Parekh}, \bibinfo{person}{Hieu Pham},
  \bibinfo{person}{Quoc Le}, \bibinfo{person}{Yun-Hsuan Sung},
  \bibinfo{person}{Zhen Li}, {and} \bibinfo{person}{Tom Duerig}.}
  \bibinfo{year}{2021}\natexlab{}.
\newblock \showarticletitle{Scaling up visual and vision-language
  representation learning with noisy text supervision}. In
  \bibinfo{booktitle}{\emph{ICML}}. \bibinfo{pages}{4904--4916}.
\newblock


\bibitem[Ju et~al\mbox{.}(2022)]%
        {ju2022prompting}
\bibfield{author}{\bibinfo{person}{Chen Ju}, \bibinfo{person}{Tengda Han},
  \bibinfo{person}{Kunhao Zheng}, \bibinfo{person}{Ya Zhang}, {and}
  \bibinfo{person}{Weidi Xie}.} \bibinfo{year}{2022}\natexlab{}.
\newblock \showarticletitle{Prompting Visual-Language Models for Efficient
  Video Understanding}. In \bibinfo{booktitle}{\emph{ECCV}}.
  \bibinfo{pages}{105--124}.
\newblock


\bibitem[Karpathy and Fei-Fei(2015)]%
        {karpathy2015deep}
\bibfield{author}{\bibinfo{person}{Andrej Karpathy} {and} \bibinfo{person}{Li
  Fei-Fei}.} \bibinfo{year}{2015}\natexlab{}.
\newblock \showarticletitle{Deep visual-semantic alignments for generating
  image descriptions}. In \bibinfo{booktitle}{\emph{CVPR}}.
  \bibinfo{pages}{3128--3137}.
\newblock


\bibitem[Laina et~al\mbox{.}(2019)]%
        {laina2019towards}
\bibfield{author}{\bibinfo{person}{Iro Laina}, \bibinfo{person}{Christian
  Rupprecht}, {and} \bibinfo{person}{Nassir Navab}.}
  \bibinfo{year}{2019}\natexlab{}.
\newblock \showarticletitle{Towards unsupervised image captioning with shared
  multimodal embeddings}. In \bibinfo{booktitle}{\emph{ICCV}}.
  \bibinfo{pages}{7414--7424}.
\newblock


\bibitem[Li et~al\mbox{.}(2023)]%
        {li2023decap}
\bibfield{author}{\bibinfo{person}{Wei Li}, \bibinfo{person}{Linchao Zhu},
  \bibinfo{person}{Longyin Wen}, {and} \bibinfo{person}{Yi Yang}.}
  \bibinfo{year}{2023}\natexlab{}.
\newblock \showarticletitle{DeCap: Decoding CLIP Latents for Zero-Shot
  Captioning via Text-Only Training}. In \bibinfo{booktitle}{\emph{ICLR}}.
\newblock


\bibitem[Li et~al\mbox{.}(2022)]%
        {li2022comprehending}
\bibfield{author}{\bibinfo{person}{Yehao Li}, \bibinfo{person}{Yingwei Pan},
  \bibinfo{person}{Ting Yao}, {and} \bibinfo{person}{Tao Mei}.}
  \bibinfo{year}{2022}\natexlab{}.
\newblock \showarticletitle{Comprehending and ordering semantics for image
  captioning}. In \bibinfo{booktitle}{\emph{CVPR}}.
  \bibinfo{pages}{17990--17999}.
\newblock


\bibitem[Liang et~al\mbox{.}(2022)]%
        {liang2022mind}
\bibfield{author}{\bibinfo{person}{Victor~Weixin Liang}, \bibinfo{person}{Yuhui
  Zhang}, \bibinfo{person}{Yongchan Kwon}, \bibinfo{person}{Serena Yeung},
  {and} \bibinfo{person}{James~Y Zou}.} \bibinfo{year}{2022}\natexlab{}.
\newblock \showarticletitle{Mind the gap: Understanding the modality gap in
  multi-modal contrastive representation learning}. In
  \bibinfo{booktitle}{\emph{NeurIPS}}. \bibinfo{pages}{17612--17625}.
\newblock


\bibitem[Lin(2004)]%
        {lin-2004-rouge}
\bibfield{author}{\bibinfo{person}{Chin-Yew Lin}.}
  \bibinfo{year}{2004}\natexlab{}.
\newblock \showarticletitle{{ROUGE}: A Package for Automatic Evaluation of
  Summaries}. In \bibinfo{booktitle}{\emph{ACL}}. \bibinfo{pages}{74--81}.
\newblock


\bibitem[Lin et~al\mbox{.}(2014)]%
        {lin2014microsoft}
\bibfield{author}{\bibinfo{person}{Tsung-Yi Lin}, \bibinfo{person}{Michael
  Maire}, \bibinfo{person}{Serge Belongie}, \bibinfo{person}{James Hays},
  \bibinfo{person}{Pietro Perona}, \bibinfo{person}{Deva Ramanan},
  \bibinfo{person}{Piotr Doll{\'a}r}, {and} \bibinfo{person}{C~Lawrence
  Zitnick}.} \bibinfo{year}{2014}\natexlab{}.
\newblock \showarticletitle{Microsoft coco: Common objects in context}. In
  \bibinfo{booktitle}{\emph{ECCV}}. \bibinfo{pages}{740--755}.
\newblock


\bibitem[Liu et~al\mbox{.}(2019a)]%
        {liu2019exploring}
\bibfield{author}{\bibinfo{person}{Fenglin Liu}, \bibinfo{person}{Meng Gao},
  \bibinfo{person}{Tianhao Zhang}, {and} \bibinfo{person}{Yuexian Zou}.}
  \bibinfo{year}{2019}\natexlab{a}.
\newblock \showarticletitle{Exploring semantic relationships for image
  captioning without parallel data}. In \bibinfo{booktitle}{\emph{ICDM}}.
  \bibinfo{pages}{439--448}.
\newblock


\bibitem[Liu et~al\mbox{.}(2019b)]%
        {liu2019roberta}
\bibfield{author}{\bibinfo{person}{Yinhan Liu}, \bibinfo{person}{Myle Ott},
  \bibinfo{person}{Naman Goyal}, \bibinfo{person}{Jingfei Du},
  \bibinfo{person}{Mandar Joshi}, \bibinfo{person}{Danqi Chen},
  \bibinfo{person}{Omer Levy}, \bibinfo{person}{Mike Lewis},
  \bibinfo{person}{Luke Zettlemoyer}, {and} \bibinfo{person}{Veselin
  Stoyanov}.} \bibinfo{year}{2019}\natexlab{b}.
\newblock \showarticletitle{Roberta: A robustly optimized bert pretraining
  approach}.
\newblock \bibinfo{journal}{\emph{arXiv preprint arXiv:1907.11692}}
  (\bibinfo{year}{2019}).
\newblock


\bibitem[Liu et~al\mbox{.}(2021a)]%
        {liu2021aggregated}
\bibfield{author}{\bibinfo{person}{Zhenguang Liu}, \bibinfo{person}{Kedi Lyu},
  \bibinfo{person}{Shuang Wu}, \bibinfo{person}{Haipeng Chen},
  \bibinfo{person}{Yanbin Hao}, {and} \bibinfo{person}{Shouling Ji}.}
  \bibinfo{year}{2021}\natexlab{a}.
\newblock \showarticletitle{Aggregated multi-gans for controlled 3d human
  motion prediction}. In \bibinfo{booktitle}{\emph{AAAI}}.
  \bibinfo{pages}{2225--2232}.
\newblock


\bibitem[Liu et~al\mbox{.}(2021b)]%
        {liu2021motion}
\bibfield{author}{\bibinfo{person}{Zhenguang Liu}, \bibinfo{person}{Pengxiang
  Su}, \bibinfo{person}{Shuang Wu}, \bibinfo{person}{Xuanjing Shen},
  \bibinfo{person}{Haipeng Chen}, \bibinfo{person}{Yanbin Hao}, {and}
  \bibinfo{person}{Meng Wang}.} \bibinfo{year}{2021}\natexlab{b}.
\newblock \showarticletitle{Motion prediction using trajectory cues}. In
  \bibinfo{booktitle}{\emph{ICCV}}. \bibinfo{pages}{13299--13308}.
\newblock


\bibitem[Liu et~al\mbox{.}(2022)]%
        {liu2022investigating}
\bibfield{author}{\bibinfo{person}{Zhenguang Liu}, \bibinfo{person}{Shuang Wu},
  \bibinfo{person}{Shuyuan Jin}, \bibinfo{person}{Shouling Ji},
  \bibinfo{person}{Qi Liu}, \bibinfo{person}{Shijian Lu}, {and}
  \bibinfo{person}{Li Cheng}.} \bibinfo{year}{2022}\natexlab{}.
\newblock \showarticletitle{Investigating pose representations and motion
  contexts modeling for 3D motion prediction}.
\newblock \bibinfo{journal}{\emph{TPAMI}} \bibinfo{volume}{45},
  \bibinfo{number}{1} (\bibinfo{year}{2022}), \bibinfo{pages}{681--697}.
\newblock


\bibitem[Loshchilov and Hutter(2018)]%
        {loshchilov2018decoupled}
\bibfield{author}{\bibinfo{person}{Ilya Loshchilov} {and}
  \bibinfo{person}{Frank Hutter}.} \bibinfo{year}{2018}\natexlab{}.
\newblock \showarticletitle{Decoupled Weight Decay Regularization}. In
  \bibinfo{booktitle}{\emph{ICLR}}.
\newblock


\bibitem[Lu et~al\mbox{.}(2017)]%
        {lu2017knowing}
\bibfield{author}{\bibinfo{person}{Jiasen Lu}, \bibinfo{person}{Caiming Xiong},
  \bibinfo{person}{Devi Parikh}, {and} \bibinfo{person}{Richard Socher}.}
  \bibinfo{year}{2017}\natexlab{}.
\newblock \showarticletitle{Knowing when to look: Adaptive attention via a
  visual sentinel for image captioning}. In \bibinfo{booktitle}{\emph{CVPR}}.
  \bibinfo{pages}{375--383}.
\newblock


\bibitem[Meng et~al\mbox{.}(2022)]%
        {meng2021object}
\bibfield{author}{\bibinfo{person}{Zihang Meng}, \bibinfo{person}{David Yang},
  \bibinfo{person}{Xuefei Cao}, \bibinfo{person}{Ashish Shah}, {and}
  \bibinfo{person}{Ser-Nam Lim}.} \bibinfo{year}{2022}\natexlab{}.
\newblock \showarticletitle{Object-Centric Unsupervised Image Captioning}. In
  \bibinfo{booktitle}{\emph{ECCV}}. \bibinfo{pages}{219--235}.
\newblock


\bibitem[Mokady et~al\mbox{.}(2021)]%
        {mokady2021clipcap}
\bibfield{author}{\bibinfo{person}{Ron Mokady}, \bibinfo{person}{Amir Hertz},
  {and} \bibinfo{person}{Amit~H Bermano}.} \bibinfo{year}{2021}\natexlab{}.
\newblock \showarticletitle{Clipcap: Clip prefix for image captioning}.
\newblock \bibinfo{journal}{\emph{arXiv preprint arXiv:2111.09734}}
  (\bibinfo{year}{2021}).
\newblock


\bibitem[Narasimhan et~al\mbox{.}(2021)]%
        {narasimhan2021clip-it}
\bibfield{author}{\bibinfo{person}{Medhini Narasimhan}, \bibinfo{person}{Anna
  Rohrbach}, {and} \bibinfo{person}{Trevor Darrell}.}
  \bibinfo{year}{2021}\natexlab{}.
\newblock \showarticletitle{CLIP-It! language-guided video summarization}. In
  \bibinfo{booktitle}{\emph{NeurIPS}}. \bibinfo{pages}{13988--14000}.
\newblock


\bibitem[Nukrai et~al\mbox{.}(2022)]%
        {david2022textonly}
\bibfield{author}{\bibinfo{person}{David Nukrai}, \bibinfo{person}{Ron Mokady},
  {and} \bibinfo{person}{Amir Globerson}.} \bibinfo{year}{2022}\natexlab{}.
\newblock \showarticletitle{Text-Only Training for Image Captioning using
  Noise-Injected CLIP}. In \bibinfo{booktitle}{\emph{EMNLP findings}}.
  \bibinfo{pages}{4055--4063}.
\newblock


\bibitem[Pan et~al\mbox{.}(2020)]%
        {pan2020x}
\bibfield{author}{\bibinfo{person}{Yingwei Pan}, \bibinfo{person}{Ting Yao},
  \bibinfo{person}{Yehao Li}, {and} \bibinfo{person}{Tao Mei}.}
  \bibinfo{year}{2020}\natexlab{}.
\newblock \showarticletitle{X-linear attention networks for image captioning}.
  In \bibinfo{booktitle}{\emph{CVPR}}. \bibinfo{pages}{10971--10980}.
\newblock


\bibitem[Papineni et~al\mbox{.}(2002)]%
        {papineni-etal-2002-bleu}
\bibfield{author}{\bibinfo{person}{Kishore Papineni}, \bibinfo{person}{Salim
  Roukos}, \bibinfo{person}{Todd Ward}, {and} \bibinfo{person}{Wei-Jing Zhu}.}
  \bibinfo{year}{2002}\natexlab{}.
\newblock \showarticletitle{{B}leu: a Method for Automatic Evaluation of
  Machine Translation}. In \bibinfo{booktitle}{\emph{ACL}}.
  \bibinfo{pages}{311--318}.
\newblock


\bibitem[Patashnik et~al\mbox{.}(2021)]%
        {patashnik2021styleclip}
\bibfield{author}{\bibinfo{person}{Or Patashnik}, \bibinfo{person}{Zongze Wu},
  \bibinfo{person}{Eli Shechtman}, \bibinfo{person}{Daniel Cohen-Or}, {and}
  \bibinfo{person}{Dani Lischinski}.} \bibinfo{year}{2021}\natexlab{}.
\newblock \showarticletitle{Styleclip: Text-driven manipulation of stylegan
  imagery}. In \bibinfo{booktitle}{\emph{ICCV}}. \bibinfo{pages}{2085--2094}.
\newblock


\bibitem[Plummer et~al\mbox{.}(2015)]%
        {plummer2015flickr30k}
\bibfield{author}{\bibinfo{person}{Bryan~A Plummer}, \bibinfo{person}{Liwei
  Wang}, \bibinfo{person}{Chris~M Cervantes}, \bibinfo{person}{Juan~C Caicedo},
  \bibinfo{person}{Julia Hockenmaier}, {and} \bibinfo{person}{Svetlana
  Lazebnik}.} \bibinfo{year}{2015}\natexlab{}.
\newblock \showarticletitle{Flickr30k entities: Collecting region-to-phrase
  correspondences for richer image-to-sentence models}. In
  \bibinfo{booktitle}{\emph{ICCV}}. \bibinfo{pages}{2641--2649}.
\newblock


\bibitem[Radford et~al\mbox{.}(2021)]%
        {radford2021learning}
\bibfield{author}{\bibinfo{person}{Alec Radford}, \bibinfo{person}{Jong~Wook
  Kim}, \bibinfo{person}{Chris Hallacy}, \bibinfo{person}{Aditya Ramesh},
  \bibinfo{person}{Gabriel Goh}, \bibinfo{person}{Sandhini Agarwal},
  \bibinfo{person}{Girish Sastry}, \bibinfo{person}{Amanda Askell},
  \bibinfo{person}{Pamela Mishkin}, \bibinfo{person}{Jack Clark},
  {et~al\mbox{.}}} \bibinfo{year}{2021}\natexlab{}.
\newblock \showarticletitle{Learning transferable visual models from natural
  language supervision}. In \bibinfo{booktitle}{\emph{ICML}}.
  \bibinfo{pages}{8748--8763}.
\newblock


\bibitem[Radford et~al\mbox{.}(2019)]%
        {radford2019language}
\bibfield{author}{\bibinfo{person}{Alec Radford}, \bibinfo{person}{Jeffrey Wu},
  \bibinfo{person}{Rewon Child}, \bibinfo{person}{David Luan},
  \bibinfo{person}{Dario Amodei}, \bibinfo{person}{Ilya Sutskever},
  {et~al\mbox{.}}} \bibinfo{year}{2019}\natexlab{}.
\newblock \showarticletitle{Language models are unsupervised multitask
  learners}.
\newblock \bibinfo{journal}{\emph{OpenAI blog}} \bibinfo{volume}{1},
  \bibinfo{number}{8} (\bibinfo{year}{2019}), \bibinfo{pages}{9}.
\newblock


\bibitem[Ramesh et~al\mbox{.}(2022)]%
        {ramesh2022hierarchical}
\bibfield{author}{\bibinfo{person}{Aditya Ramesh}, \bibinfo{person}{Prafulla
  Dhariwal}, \bibinfo{person}{Alex Nichol}, \bibinfo{person}{Casey Chu}, {and}
  \bibinfo{person}{Mark Chen}.} \bibinfo{year}{2022}\natexlab{}.
\newblock \showarticletitle{Hierarchical text-conditional image generation with
  clip latents}.
\newblock \bibinfo{journal}{\emph{arXiv preprint arXiv:2204.06125}}
  (\bibinfo{year}{2022}).
\newblock


\bibitem[Rennie et~al\mbox{.}(2017)]%
        {rennie2017self}
\bibfield{author}{\bibinfo{person}{Steven~J Rennie}, \bibinfo{person}{Etienne
  Marcheret}, \bibinfo{person}{Youssef Mroueh}, \bibinfo{person}{Jerret Ross},
  {and} \bibinfo{person}{Vaibhava Goel}.} \bibinfo{year}{2017}\natexlab{}.
\newblock \showarticletitle{Self-critical sequence training for image
  captioning}. In \bibinfo{booktitle}{\emph{CVPR}}.
  \bibinfo{pages}{7008--7024}.
\newblock


\bibitem[Sharma et~al\mbox{.}(2018)]%
        {sharma2018conceptual}
\bibfield{author}{\bibinfo{person}{Piyush Sharma}, \bibinfo{person}{Nan Ding},
  \bibinfo{person}{Sebastian Goodman}, {and} \bibinfo{person}{Radu Soricut}.}
  \bibinfo{year}{2018}\natexlab{}.
\newblock \showarticletitle{Conceptual captions: A cleaned, hypernymed, image
  alt-text dataset for automatic image captioning}. In
  \bibinfo{booktitle}{\emph{ACL}}. \bibinfo{pages}{2556--2565}.
\newblock


\bibitem[Song et~al\mbox{.}(2022a)]%
        {song2022clip}
\bibfield{author}{\bibinfo{person}{Haoyu Song}, \bibinfo{person}{Li Dong},
  \bibinfo{person}{Weinan Zhang}, \bibinfo{person}{Ting Liu}, {and}
  \bibinfo{person}{Furu Wei}.} \bibinfo{year}{2022}\natexlab{a}.
\newblock \showarticletitle{CLIP Models are Few-Shot Learners: Empirical
  Studies on VQA and Visual Entailment}. In \bibinfo{booktitle}{\emph{ACL}}.
  \bibinfo{pages}{6088--6100}.
\newblock


\bibitem[Song et~al\mbox{.}(2022b)]%
        {song2022memorial}
\bibfield{author}{\bibinfo{person}{Peipei Song}, \bibinfo{person}{Dan Guo},
  \bibinfo{person}{Jinxing Zhou}, \bibinfo{person}{Mingliang Xu}, {and}
  \bibinfo{person}{Meng Wang}.} \bibinfo{year}{2022}\natexlab{b}.
\newblock \showarticletitle{Memorial GAN With Joint Semantic Optimization for
  Unpaired Image Captioning}.
\newblock \bibinfo{journal}{\emph{TCyber}} (\bibinfo{year}{2022}).
\newblock


\bibitem[Song et~al\mbox{.}(2019)]%
        {song2019unpaired}
\bibfield{author}{\bibinfo{person}{Yuqing Song}, \bibinfo{person}{Shizhe Chen},
  \bibinfo{person}{Yida Zhao}, {and} \bibinfo{person}{Qin Jin}.}
  \bibinfo{year}{2019}\natexlab{}.
\newblock \showarticletitle{Unpaired cross-lingual image caption generation
  with self-supervised rewards}. In \bibinfo{booktitle}{\emph{ACM MM}}.
  \bibinfo{pages}{784--792}.
\newblock


\bibitem[Su et~al\mbox{.}(2022)]%
        {su2022language}
\bibfield{author}{\bibinfo{person}{Yixuan Su}, \bibinfo{person}{Tian Lan},
  \bibinfo{person}{Yahui Liu}, \bibinfo{person}{Fangyu Liu},
  \bibinfo{person}{Dani Yogatama}, \bibinfo{person}{Yan Wang},
  \bibinfo{person}{Lingpeng Kong}, {and} \bibinfo{person}{Nigel Collier}.}
  \bibinfo{year}{2022}\natexlab{}.
\newblock \showarticletitle{Language models can see: plugging visual controls
  in text generation}.
\newblock \bibinfo{journal}{\emph{arXiv preprint arXiv:2205.02655}}
  (\bibinfo{year}{2022}).
\newblock


\bibitem[Sutton and Barto(2018)]%
        {sutton2018reinforcement}
\bibfield{author}{\bibinfo{person}{Richard~S Sutton} {and}
  \bibinfo{person}{Andrew~G Barto}.} \bibinfo{year}{2018}\natexlab{}.
\newblock \bibinfo{booktitle}{\emph{Reinforcement learning: An introduction}}.
\newblock \bibinfo{publisher}{MIT press}.
\newblock


\bibitem[Tang et~al\mbox{.}(2021)]%
        {tang2021clip4caption}
\bibfield{author}{\bibinfo{person}{Mingkang Tang}, \bibinfo{person}{Zhanyu
  Wang}, \bibinfo{person}{Zhenhua Liu}, \bibinfo{person}{Fengyun Rao},
  \bibinfo{person}{Dian Li}, {and} \bibinfo{person}{Xiu Li}.}
  \bibinfo{year}{2021}\natexlab{}.
\newblock \showarticletitle{Clip4caption: Clip for video caption}. In
  \bibinfo{booktitle}{\emph{ACM MM}}. \bibinfo{pages}{4858--4862}.
\newblock


\bibitem[Tewel et~al\mbox{.}(2022)]%
        {tewel2022zerocap}
\bibfield{author}{\bibinfo{person}{Yoad Tewel}, \bibinfo{person}{Yoav Shalev},
  \bibinfo{person}{Idan Schwartz}, {and} \bibinfo{person}{Lior Wolf}.}
  \bibinfo{year}{2022}\natexlab{}.
\newblock \showarticletitle{ZeroCap: Zero-Shot Image-to-Text Generation for
  Visual-Semantic Arithmetic}. In \bibinfo{booktitle}{\emph{CVPR}}.
  \bibinfo{pages}{17918--17928}.
\newblock


\bibitem[Vaswani et~al\mbox{.}(2017)]%
        {vaswani2017attention}
\bibfield{author}{\bibinfo{person}{Ashish Vaswani}, \bibinfo{person}{Noam
  Shazeer}, \bibinfo{person}{Niki Parmar}, \bibinfo{person}{Jakob Uszkoreit},
  \bibinfo{person}{Llion Jones}, \bibinfo{person}{Aidan~N Gomez},
  \bibinfo{person}{{\L}ukasz Kaiser}, {and} \bibinfo{person}{Illia
  Polosukhin}.} \bibinfo{year}{2017}\natexlab{}.
\newblock \showarticletitle{Attention is all you need}. In
  \bibinfo{booktitle}{\emph{NeurIPS}}. \bibinfo{pages}{5998--6008}.
\newblock


\bibitem[Vedantam et~al\mbox{.}(2015)]%
        {vedantam2015cider}
\bibfield{author}{\bibinfo{person}{Ramakrishna Vedantam}, \bibinfo{person}{C
  Lawrence~Zitnick}, {and} \bibinfo{person}{Devi Parikh}.}
  \bibinfo{year}{2015}\natexlab{}.
\newblock \showarticletitle{Cider: Consensus-based image description
  evaluation}. In \bibinfo{booktitle}{\emph{CVPR}}.
  \bibinfo{pages}{4566--4575}.
\newblock


\bibitem[Vinyals et~al\mbox{.}(2016)]%
        {vinyals2016show}
\bibfield{author}{\bibinfo{person}{Oriol Vinyals}, \bibinfo{person}{Alexander
  Toshev}, \bibinfo{person}{Samy Bengio}, {and} \bibinfo{person}{Dumitru
  Erhan}.} \bibinfo{year}{2016}\natexlab{}.
\newblock \showarticletitle{Show and tell: Lessons learned from the 2015 mscoco
  image captioning challenge}.
\newblock \bibinfo{journal}{\emph{TPAMI}} \bibinfo{volume}{39},
  \bibinfo{number}{4} (\bibinfo{year}{2016}), \bibinfo{pages}{652--663}.
\newblock


\bibitem[Wang et~al\mbox{.}(2023)]%
        {wang2023generative}
\bibfield{author}{\bibinfo{person}{Wenjie Wang}, \bibinfo{person}{Xinyu Lin},
  \bibinfo{person}{Fuli Feng}, \bibinfo{person}{Xiangnan He}, {and}
  \bibinfo{person}{Tat-Seng Chua}.} \bibinfo{year}{2023}\natexlab{}.
\newblock \showarticletitle{Generative recommendation: Towards next-generation
  recommender paradigm}.
\newblock \bibinfo{journal}{\emph{arXiv preprint arXiv:2304.03516}}
  (\bibinfo{year}{2023}).
\newblock


\bibitem[Wang et~al\mbox{.}(2022a)]%
        {wang2022cris}
\bibfield{author}{\bibinfo{person}{Zhaoqing Wang}, \bibinfo{person}{Yu Lu},
  \bibinfo{person}{Qiang Li}, \bibinfo{person}{Xunqiang Tao},
  \bibinfo{person}{Yandong Guo}, \bibinfo{person}{Mingming Gong}, {and}
  \bibinfo{person}{Tongliang Liu}.} \bibinfo{year}{2022}\natexlab{a}.
\newblock \showarticletitle{Cris: Clip-driven referring image segmentation}. In
  \bibinfo{booktitle}{\emph{CVPR}}. \bibinfo{pages}{11686--11695}.
\newblock


\bibitem[Wang et~al\mbox{.}(2022b)]%
        {wangsimvlm}
\bibfield{author}{\bibinfo{person}{Zirui Wang}, \bibinfo{person}{Jiahui Yu},
  \bibinfo{person}{Adams~Wei Yu}, \bibinfo{person}{Zihang Dai},
  \bibinfo{person}{Yulia Tsvetkov}, {and} \bibinfo{person}{Yuan Cao}.}
  \bibinfo{year}{2022}\natexlab{b}.
\newblock \showarticletitle{SimVLM: Simple Visual Language Model Pretraining
  with Weak Supervision}. In \bibinfo{booktitle}{\emph{ICLR}}.
\newblock


\bibitem[Xu et~al\mbox{.}(2016)]%
        {xu2016msr}
\bibfield{author}{\bibinfo{person}{Jun Xu}, \bibinfo{person}{Tao Mei},
  \bibinfo{person}{Ting Yao}, {and} \bibinfo{person}{Yong Rui}.}
  \bibinfo{year}{2016}\natexlab{}.
\newblock \showarticletitle{Msr-vtt: A large video description dataset for
  bridging video and language}. In \bibinfo{booktitle}{\emph{CVPR}}.
  \bibinfo{pages}{5288--5296}.
\newblock


\bibitem[Xu et~al\mbox{.}(2015)]%
        {xu2015show}
\bibfield{author}{\bibinfo{person}{Kelvin Xu}, \bibinfo{person}{Jimmy Ba},
  \bibinfo{person}{Ryan Kiros}, \bibinfo{person}{Kyunghyun Cho},
  \bibinfo{person}{Aaron Courville}, \bibinfo{person}{Ruslan Salakhudinov},
  \bibinfo{person}{Rich Zemel}, {and} \bibinfo{person}{Yoshua Bengio}.}
  \bibinfo{year}{2015}\natexlab{}.
\newblock \showarticletitle{Show, attend and tell: Neural image caption
  generation with visual attention}. In \bibinfo{booktitle}{\emph{ICML}}.
  \bibinfo{pages}{2048--2057}.
\newblock


\bibitem[Yao et~al\mbox{.}(2018)]%
        {yao2018exploring}
\bibfield{author}{\bibinfo{person}{Ting Yao}, \bibinfo{person}{Yingwei Pan},
  \bibinfo{person}{Yehao Li}, {and} \bibinfo{person}{Tao Mei}.}
  \bibinfo{year}{2018}\natexlab{}.
\newblock \showarticletitle{Exploring visual relationship for image
  captioning}. In \bibinfo{booktitle}{\emph{ECCV}}. \bibinfo{pages}{684--699}.
\newblock


\bibitem[Yu et~al\mbox{.}(2017)]%
        {yu2017seqgan}
\bibfield{author}{\bibinfo{person}{Lantao Yu}, \bibinfo{person}{Weinan Zhang},
  \bibinfo{person}{Jun Wang}, {and} \bibinfo{person}{Yong Yu}.}
  \bibinfo{year}{2017}\natexlab{}.
\newblock \showarticletitle{Seqgan: Sequence generative adversarial nets with
  policy gradient}. In \bibinfo{booktitle}{\emph{AAAI}}.
\newblock


\bibitem[Yu et~al\mbox{.}(2022)]%
        {yu2022multimodal}
\bibfield{author}{\bibinfo{person}{Youngjae Yu}, \bibinfo{person}{Jiwan Chung},
  \bibinfo{person}{Heeseung Yun}, \bibinfo{person}{Jack Hessel},
  \bibinfo{person}{JaeSung Park}, \bibinfo{person}{Ximing Lu},
  \bibinfo{person}{Prithviraj Ammanabrolu}, \bibinfo{person}{Rowan Zellers},
  \bibinfo{person}{Ronan~Le Bras}, \bibinfo{person}{Gunhee Kim},
  {et~al\mbox{.}}} \bibinfo{year}{2022}\natexlab{}.
\newblock \showarticletitle{Multimodal Knowledge Alignment with Reinforcement
  Learning}.
\newblock \bibinfo{journal}{\emph{arXiv preprint arXiv:2205.12630}}
  (\bibinfo{year}{2022}).
\newblock


\bibitem[Zellers et~al\mbox{.}(2018)]%
        {zellers2018neural}
\bibfield{author}{\bibinfo{person}{Rowan Zellers}, \bibinfo{person}{Mark
  Yatskar}, \bibinfo{person}{Sam Thomson}, {and} \bibinfo{person}{Yejin Choi}.}
  \bibinfo{year}{2018}\natexlab{}.
\newblock \showarticletitle{Neural motifs: Scene graph parsing with global
  context}. In \bibinfo{booktitle}{\emph{CVPR}}. \bibinfo{pages}{5831--5840}.
\newblock


\bibitem[Zhou et~al\mbox{.}(2021)]%
        {zhou2021triple}
\bibfield{author}{\bibinfo{person}{Yucheng Zhou}, \bibinfo{person}{Wei Tao},
  {and} \bibinfo{person}{Wenqiang Zhang}.} \bibinfo{year}{2021}\natexlab{}.
\newblock \showarticletitle{Triple sequence generative adversarial nets for
  unsupervised image captioning}. In \bibinfo{booktitle}{\emph{ICASSP}}.
  \bibinfo{pages}{7598--7602}.
\newblock


\bibitem[Zhu and Ngo(2020)]%
        {zhu2020cookgan}
\bibfield{author}{\bibinfo{person}{Bin Zhu} {and} \bibinfo{person}{Chong-Wah
  Ngo}.} \bibinfo{year}{2020}\natexlab{}.
\newblock \showarticletitle{CookGAN: Causality based text-to-image synthesis}.
  In \bibinfo{booktitle}{\emph{CVPR}}. \bibinfo{pages}{5519--5527}.
\newblock


\bibitem[Zhu et~al\mbox{.}(2019)]%
        {zhu2019r2gan}
\bibfield{author}{\bibinfo{person}{Bin Zhu}, \bibinfo{person}{Chong-Wah Ngo},
  \bibinfo{person}{Jingjing Chen}, {and} \bibinfo{person}{Yanbin Hao}.}
  \bibinfo{year}{2019}\natexlab{}.
\newblock \showarticletitle{R2gan: Cross-modal recipe retrieval with generative
  adversarial network}. In \bibinfo{booktitle}{\emph{CVPR}}.
  \bibinfo{pages}{11477--11486}.
\newblock


\bibitem[Zhu et~al\mbox{.}(2022)]%
        {zhu2022unpaired}
\bibfield{author}{\bibinfo{person}{Peipei Zhu}, \bibinfo{person}{Xiao Wang},
  \bibinfo{person}{Yong Luo}, \bibinfo{person}{Zhenglong Sun},
  \bibinfo{person}{Wei-Shi Zheng}, \bibinfo{person}{Yaowei Wang}, {and}
  \bibinfo{person}{Changwen Chen}.} \bibinfo{year}{2022}\natexlab{}.
\newblock \showarticletitle{Unpaired Image Captioning by Image-level
  Weakly-Supervised Visual Concept Recognition}.
\newblock \bibinfo{journal}{\emph{TMM}} (\bibinfo{year}{2022}),
  \bibinfo{pages}{1--15}.
\newblock


\bibitem[Zhu et~al\mbox{.}(2023)]%
        {zhu2023prompt}
\bibfield{author}{\bibinfo{person}{Peipei Zhu}, \bibinfo{person}{Xiao Wang},
  \bibinfo{person}{Lin Zhu}, \bibinfo{person}{Zhenglong Sun},
  \bibinfo{person}{Wei-Shi Zheng}, \bibinfo{person}{Yaowei Wang}, {and}
  \bibinfo{person}{Changwen Chen}.} \bibinfo{year}{2023}\natexlab{}.
\newblock \showarticletitle{Prompt-based learning for unpaired image
  captioning}.
\newblock \bibinfo{journal}{\emph{TMM}} (\bibinfo{year}{2023}),
  \bibinfo{pages}{1--15}.
\newblock


\end{thebibliography}
